\documentclass{article}
\usepackage{hyperref}
\usepackage[accepted]{icml2024_arxiv}
\usepackage{microtype}
\usepackage{graphicx}
\usepackage{subfigure}
\usepackage{booktabs} % for professional tables
\usepackage{hyperref}
\usepackage{algorithm}
\usepackage{algpseudocode}

\usepackage{colortbl}
\usepackage{lipsum}

\usepackage{amsmath}
\usepackage{amssymb}
\usepackage{mathtools}
\usepackage{amsthm}

\usepackage[capitalize,noabbrev]{cleveref}
\usepackage{colortbl}
\usepackage{nicematrix}
\usepackage[inline, shortlabels]{enumitem}
\usepackage{multirow}

\usepackage{amsthm}
\usepackage{amsmath,amsfonts,bm}
\usepackage{nicefrac}
\usepackage{array}
\usepackage{wrapfig}

\usepackage{hyperref}
\usepackage{url}
\usepackage{tabu}

\usepackage{pifont} 
\newcommand{\cmark}{\ding{51}}%
\newcommand{\xmark}{\ding{55}}%

\usepackage[listings]{tcolorbox}
\tcbuselibrary{listings,theorems}

\newtcolorbox{mybox}[1]{colback=green2!5,colframe=blue1,fonttitle=\bfseries, left=.0in, right=.0in,bottom=0in, top=0in, title=#1}

\definecolor{bluex}{rgb}{0.27, 0.42, 0.81}
\definecolor{purplex}{HTML}{9564bf}

\definecolor{bluex}{rgb}{0.27, 0.42, 0.81}
\definecolor{purplex}{HTML}{9564bf}
\definecolor{red3}{HTML}{C52A20}
\definecolor{red2}{HTML}{B36A6F}
\definecolor{red1}{HTML}{FFb5b5}
\definecolor{purple}{HTML}{B36A6F}
\definecolor{darkyellow}{HTML}{D5BA82}
\definecolor{blue1}{HTML}{508AB2}
\definecolor{blue2}{HTML}{C4E4E3}
\definecolor{blue3}{HTML}{1663A9}
\definecolor{green1}{HTML}{A1D0C7}
\definecolor{green2}{HTML}{BFF6BA}
\definecolor{green3}{HTML}{028100}
\definecolor{teal}{HTML}{508AB2}
\definecolor{orange}{HTML}{f0b37b}
\definecolor{Gray}{gray}{0.94}
\definecolor{rev}{HTML}{000000}
\theoremstyle{plain}

\theoremstyle{definition}

\theoremstyle{remark}

\newcommand{\bR}{\mathbb{R}}

\newcommand{\STAB}[1]{\begin{tabular}{@{}c@{}}#1\end{tabular}}

\newcommand{\hA}{\mathcal{A}}

\newcommand{\vu}{{\bf u}}
\newcommand{\vv}{{\bf v}}

\newcommand{\vx}{{\bf x}}

\newcommand{\vA}{{\bf A}}
\newcommand{\vB}{{\bf B}}

\newcommand{\vU}{{\bf U}}
\newcommand{\vV}{{\bf V}}

\newcommand{\vtheta}{{\boldsymbol \theta}}

\newcommand{\vSigma}{{\boldsymbol \Sigma}}

\usepackage{dblfloatfix}
\begin{document}
	
	\twocolumn[
	\icmltitle{BYOM: Building Your Own Multi-Task Model For Free}
	
	\icmlsetsymbol{equal}{*}
	
	\begin{icmlauthorlist}
		
		\icmlauthor{Weisen Jiang}{equal,x1,x2}
		\icmlauthor{Baijiong Lin}{equal,x5}
		\icmlauthor{Han Shi}{x3}
		\icmlauthor{Yu Zhang}{x1,x4}
		\icmlauthor{Zhenguo Li}{x3}
		\icmlauthor{James T. Kwok}{x2}
	\end{icmlauthorlist}
	
	\icmlaffiliation{x1}{Southern University of Science and Technology}
	\icmlaffiliation{x2}{Hong Kong University of Science and Technology}
	\icmlaffiliation{x3}{Huawei Noah’s Ark Lab}
	\icmlaffiliation{x4}{Peng Cheng Laboratory}
	\icmlaffiliation{x5}{Hong Kong University of Science and Technology (Guangzhou)}
	
	\icmlcorrespondingauthor{Yu Zhang}{yu.zhang.ust@gmail.com}
	
	\icmlkeywords{Deep Learning, Merging models}
	\vskip 0.3in
	]
	\printAffiliationsAndNotice{\icmlEqualContribution} 
	
	%%%%%%%%%%%%%%%%%%%%%%%%%%%%%%%%%
	%% Section: Anstract
	%%%%%%%%%%%%%%%%%%%%%%%%%%%%%%%%%
	
	\begin{abstract} 
		Recently, various merging methods have been proposed to build a multi-task model from task-specific finetuned models without retraining. However, existing methods suffer from a large performance deterioration compared to using multiple task-specific models. In this paper, we propose to inject task-specific knowledge into the merged model and design two parameter-efficient approaches (BYOM-FFT and BYOM-LoRA) to \textbf{B}uild \textbf{Y}our \textbf{O}wn \textbf{M}ulti-task model. BYOM-FFT is for merging fully finetuned models, while BYOM-LoRA is for LoRA-finetuned models. Both methods are data-free and computation-efficient. Extensive experiments on computer vision and natural language processing tasks show that the proposed BYOM methods outperform existing merging methods by a large margin. Moreover, BYOM-FFT is general and can be integrated into existing merging methods to further boost performance.
	\end{abstract}
	
	%%%%%%%%%%%%%%%%%%%%%%%%%%%%%%%%%
	%% Section: Introduction
	%%%%%%%%%%%%%%%%%%%%%%%%%%%%%%%%%
	\section{Introduction}
	\label{sec:intro}	
	In recent years,
	large-scale foundation models pre-trained on massive data have proven effective in transferring to downstream tasks~\citep{chen2022meta, min2022metaicl,yuan2023scaling, ruiz2023dreambooth}.
	Various models are available on Hugging Face~\citep{wolf2020transformers},
	e.g.,
	\textit{ResNet}~\citep{he2016deep}, \textit{ViT}~\citep{dosovitskiy2021an},  \textit{CLIP}~\citep{radford2021learning}, and diffusion models~\citep{ho2020denoising, rombach2022high} for computer vision (CV);
	\textit{T5}~\citep{Colin2020t5}, 
	\textit{GPT-2}~\citep{gpt2}, and \textit{LLaMA}~\citep{touvron2023llama1, touvron2023llama} models for natural language processing (NLP).
	Practitioners specialize a
	pre-trained model 
	to a task-specific model by either fully or 
	parameter-efficient finetuning on the task data~\citep{houlsby2019parameter, hu2021lora,lester2021power, jiang2023effective, yu2023metamath}.
	Many task-specific finetuned models 
	are published online
	for public use.
	By 2024, more than $120,000$ models have been released on Hugging Face. 
	
	\begin{table}[!t]
		\centering
		% \vskip -.2in
		\caption{Comparison between BYOM and merging methods in terms of Data-Efficiency (DE), Computation-Efficiency (CE),   Negligible Deterioration (ND) in performance compared to the single-task method.}
		\label{tab:compare-method}
		\label{table:}
		\resizebox{1\linewidth}{!}{
			\begin{tabular}{lccc}
				\toprule
				& DE & CE  & ND \\
				\midrule
				Fisher-Merging~\citep{matena2022merging} &\xmark & \xmark    &\xmark \\
				RegMean~\citep{jin2023dataless} &\xmark & \cmark    &\xmark \\
				AdaMerging~\citep{yang2023adamerging} &\xmark & \xmark     &\xmark \\
				UncertaintyMerging~\citep{daheim2023model} &\xmark & \xmark  &\xmark \\
				\midrule
				Task-Arithmetic~\citep{ilharco2023editing} & \cmark & \cmark   &\xmark \\
				TIES-Merging~\citep{yadav2023resolving} &\cmark & \cmark  &\xmark \\ 
				\rowcolor{Gray}
				BYOM (\textbf{ours}) &\cmark & \cmark    &\cmark\\
				\bottomrule
			\end{tabular}
		} 
		\vskip -.2in
	\end{table}
	
	However, in real-world applications,
	we usually deal with many tasks simultaneously~\citep{dong2015multi, siam2018modnet, Colin2020t5},
	e.g., building a powerful multilingual model (English, Chinese, Japanese, Korean).
	Using a task-specific finetuned model for each task
	is effective but costly in storing and serving.
	On the other hand,
	training a multi-task learning (MTL) model \citep{zhang2021survey} (in a parameter-efficient manner)
	is promising but requires the availability of all task-specific training data and expensive computations.
	Numerous companies and users are open to sharing their model checkpoints online, but they are reluctant to share the training data due to privacy concerns or commercial competition.
	
	\begin{figure*}[!h]
		\centering
		\subfigure[\label{fig:vitb32_bubble_sparse}\textit{ViT-B/32}.]{\includegraphics[width=0.33\textwidth]{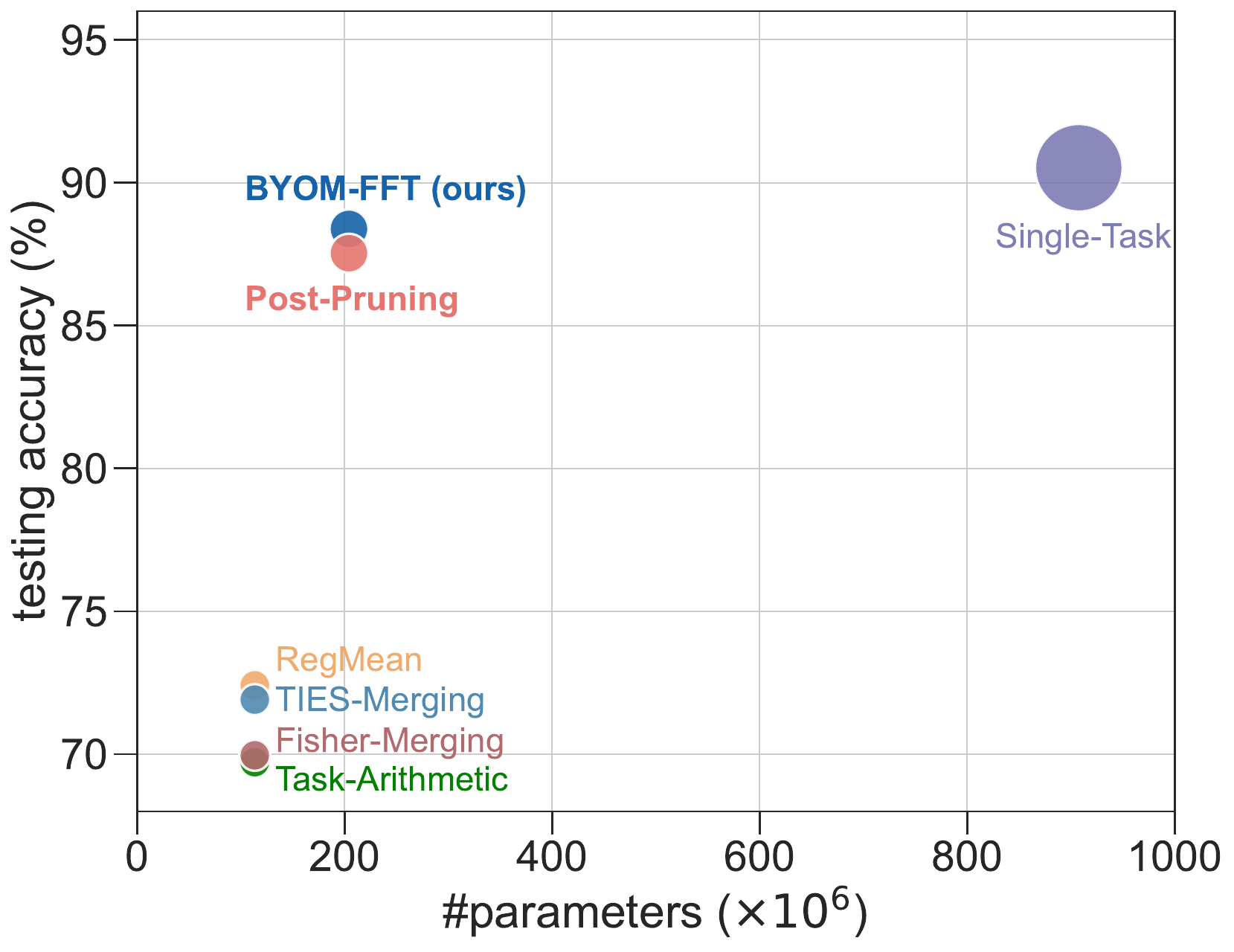}}
		\subfigure[\label{fig:vitb16_bubble_sparse}\textit{ViT-B/16}.]{\includegraphics[width=0.33\textwidth]{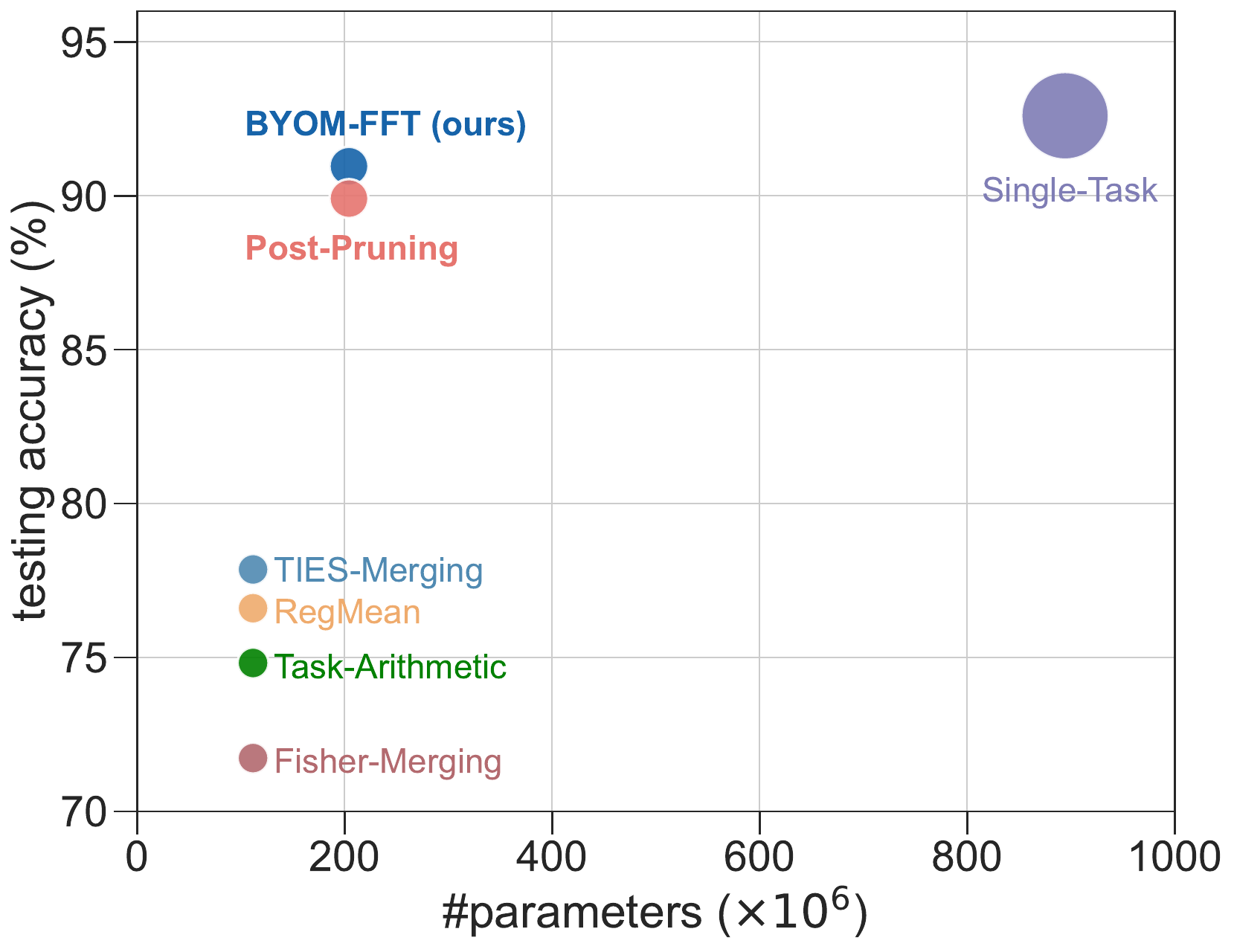}} \subfigure[\label{fig:vitL14_bubble_sparse}\textit{ViT-L/14}.]{\includegraphics[width=0.33\textwidth]{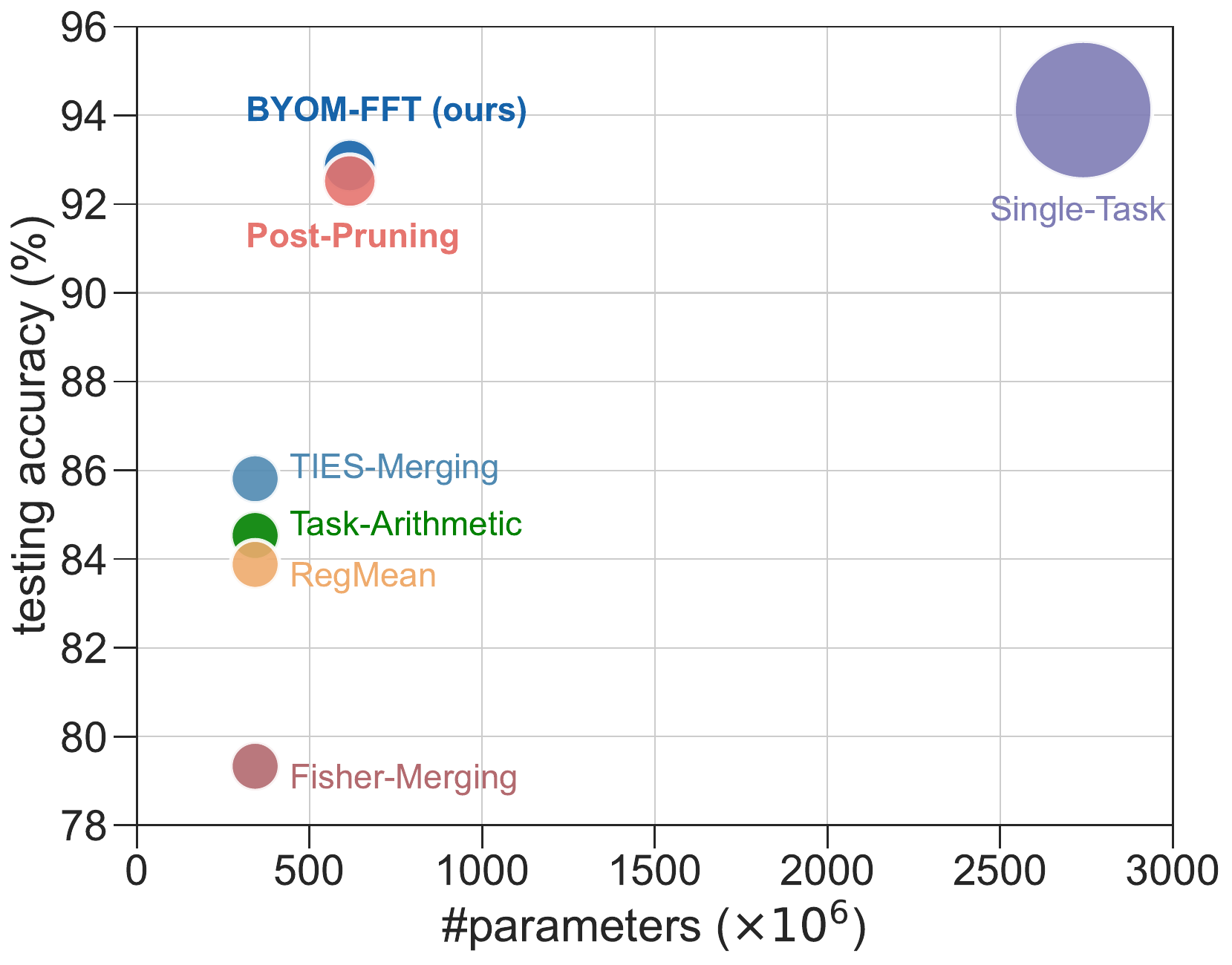}} 
		\vskip -.15in
		\caption{Testing accuracy (averaged over eight tasks) of methods merging fully finetuned models.
			% Post-Pruning and BYOM-FFT keep top-$10\%$ values.
		}
		\label{fig:vitb32-sparse-bubble}
		\vskip -.15in
	\end{figure*}
	
	To mitigate the issues of data unavailability and expensive computations,
	various merging methods have been proposed to 
	build a multi-task model from
	task-specific finetuned models without training.
	Task-Arithmetic \citep{ilharco2023editing} 
	proposes a simple merging method by averaging task vectors,
	while TIES-Merging~\citep{yadav2023resolving} 
	performs pruning before merging.
	Task-Arithmetic and TIES-Merging
	achieve data-efficiency and computation-efficiency but sacrifice performance.
	Other methods like Fisher-Merging~\citep{matena2022merging}, RegMean~\citep{jin2023dataless}, AdaMerging~\citep{yang2023adamerging}, UncertantyMerging~\citep{daheim2023model} 
	still require some data and expensive computations during the merging process,
	as summarized in Table \ref{tab:compare-method}.
	Figure \ref{fig:vitb32-sparse-bubble}
	shows the testing accuracy (averaged over eight tasks) when merging finetuned \textit{ViT} models~\citep{dosovitskiy2021an, radford2021learning},
	demonstrating
	a large gap between the performance of
	existing merging methods
	and that of the single-task method (denoted Single-Task).

	\begin{figure*}[!h]
		\centering
		\begin{minipage}{0.34\linewidth}
			\centering
			% \vskip .1in
			\includegraphics[width=0.99\linewidth]{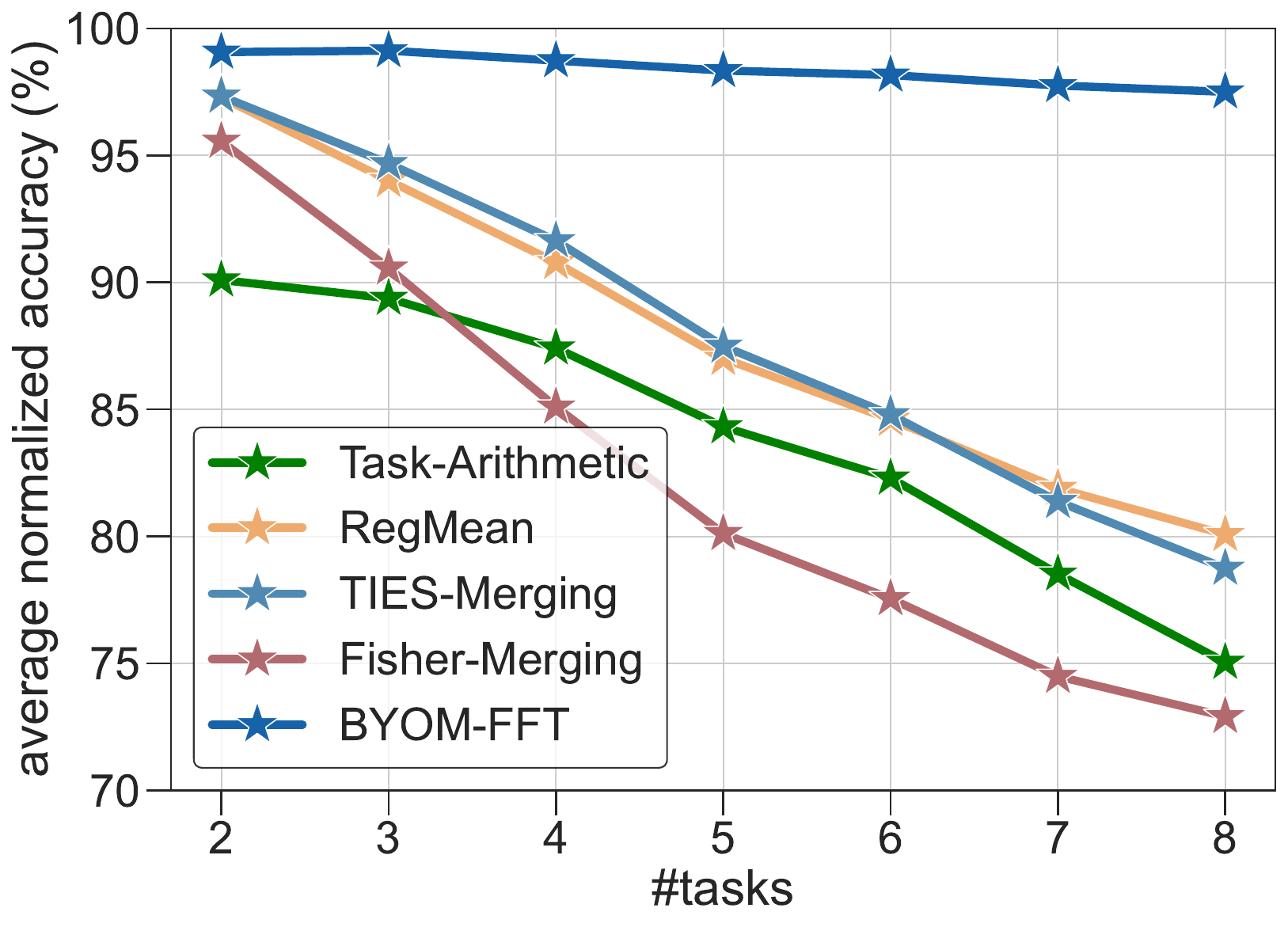} \vskip -.15in
			\caption{Relative performance with the number of tasks in merging task-specific models fully finetuned from \textit{ViT-B/32}.} 
			\label{fig:fft-num-task-relative-acc}
			\vspace{-1em}
		\end{minipage}  \;\;\;
		\begin{minipage}{0.3\linewidth}
			\centering
			\includegraphics[width=0.99\linewidth]{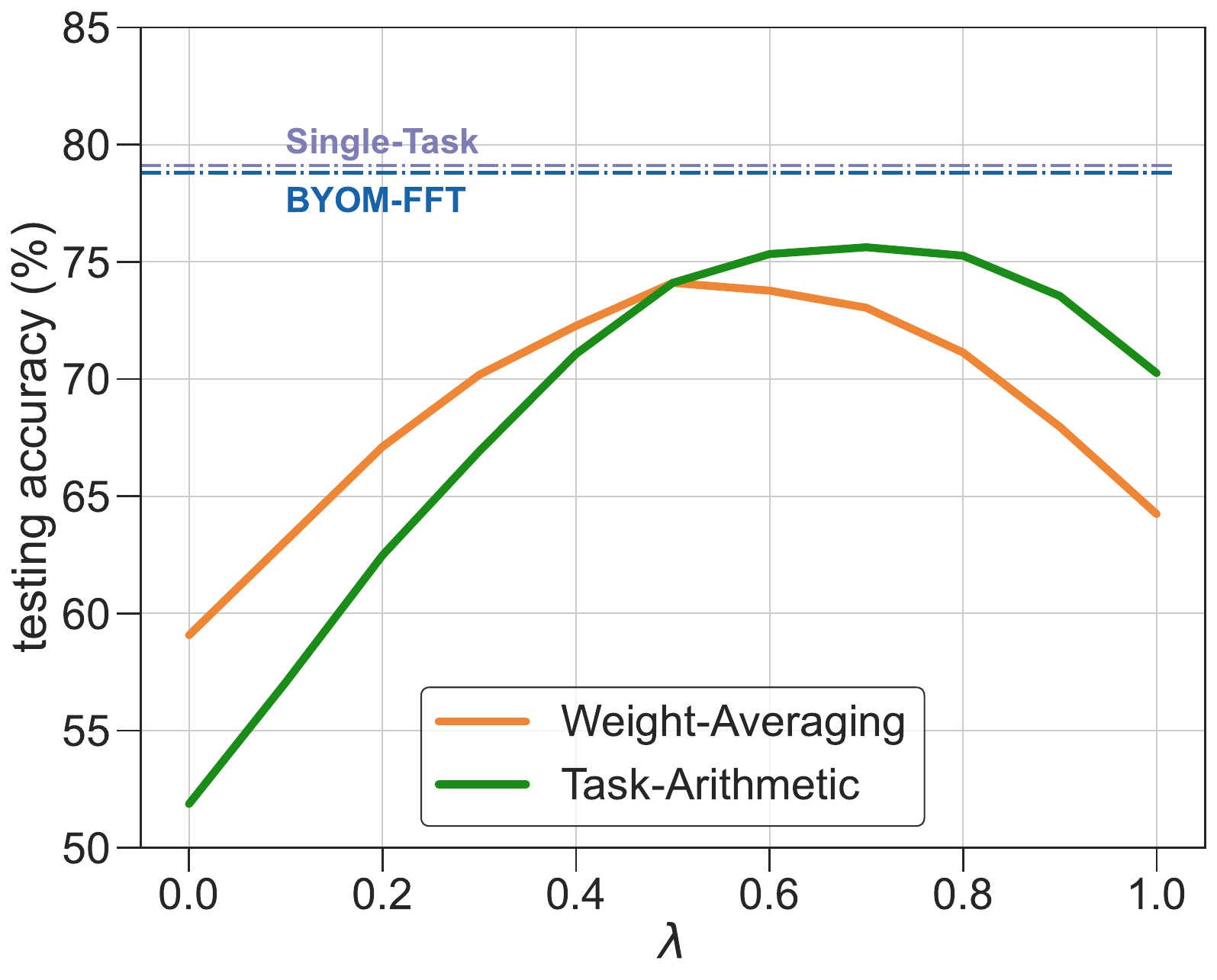}\vskip -.15in
			\caption{Testing accuracy (averaged over two tasks) of merging methods on  two \textit{dissimilar} tasks \textit{DTD} and \textit{Cars}.} 
			\label{fig:dis-sim-task}
			\vspace{-1em}
		\end{minipage}  \;\;\;
		\begin{minipage}{0.3\linewidth}
			\centering
			\includegraphics[width=0.99\linewidth]{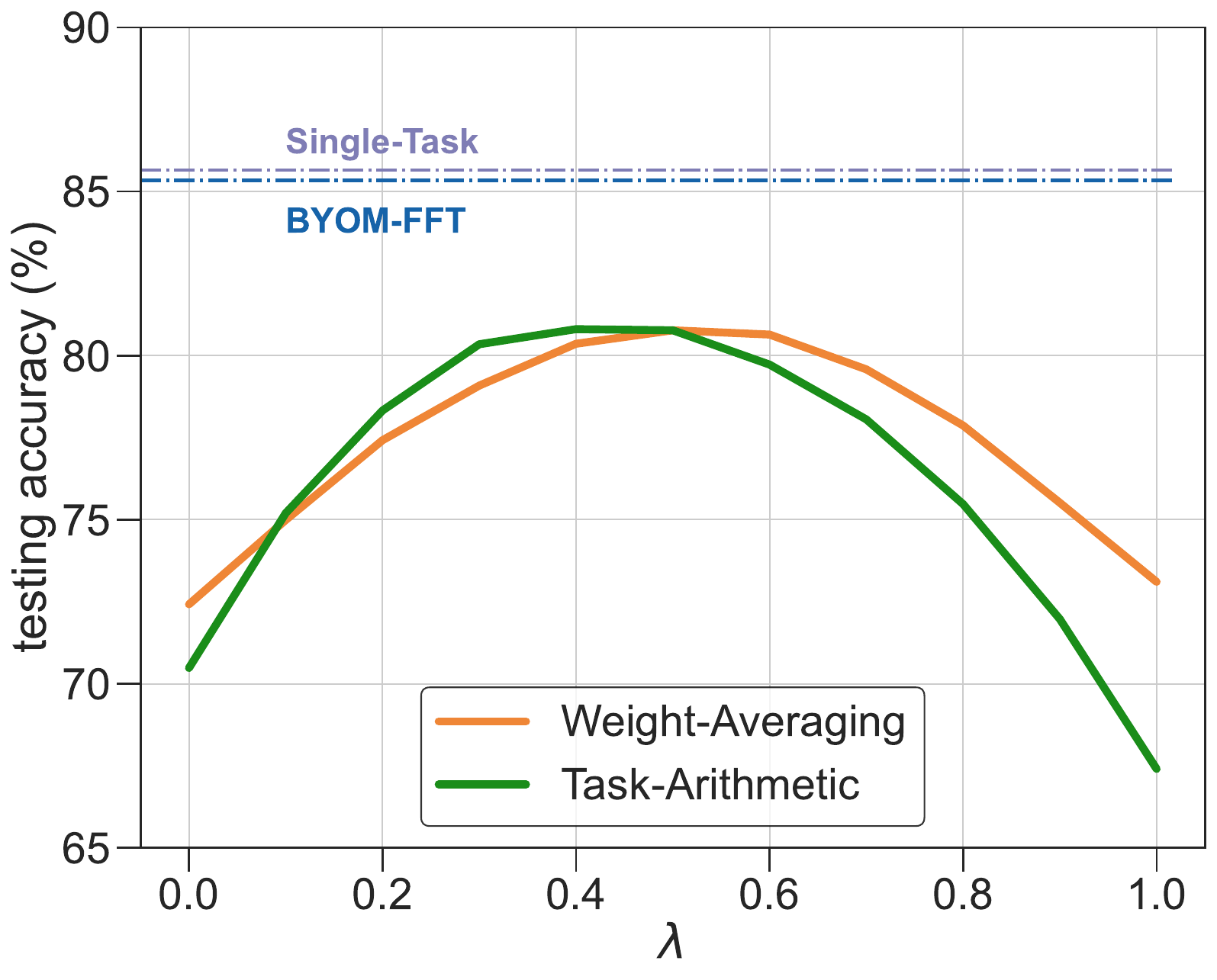}\vskip -.15in
			\caption{Testing accuracy (averaged over two tasks) of merging methods on two \textit{similar} tasks split from \textit{Cars}.} 
			\label{fig:sim-task}
			\vspace{-1em}
		\end{minipage}
	\end{figure*}
	
	In this paper,
	we propose injecting task-specific knowledge into a shared model
	to \textbf{B}uild \textbf{Y}our \textbf{O}wn \textbf{M}ulti-task model (BYOM).
	For reusing fully finetuned (FFT) models, 
	task-specific knowledge is compressed into a sparse vector.
	We propose BYOM-FFT to build a multi-task model based on the shared model and sparse vectors.
	For reusing LoRA-finetuned models (with rank $r$),
	task-specific knowledge is compressed into lower-rank matrices of rank $q$ ($q\ll r$) by singular value decomposition.
	We propose BYOM-LoRA to build a multi-task model based on the shared model and rank-$q$ matrices.
	We conduct extensive experiments on CV and NLP tasks using various network architectures 
	to demonstrate the superiority of the proposed BYOM.
	
	Our contributions are  summarized as follows:
	\begin{enumerate*}[(i), series = tobecont, itemjoin = \quad]
		\item 
		We propose BYOM-FFT and BYOM-LoRA 
		to build a multi-task model from fully and LoRA finetuned task-specific models, respectively.
		Both methods are \textbf{parameter-efficient}:
		Task-specific knowledge is compressed into 
		a sparse vector/matrix with few parameters.
		\item BYOM-FFT and BYOM-LoRA 
		are \textbf{data-free} and \textbf{training-free} in the merging process.
		\item 
		Extensive experiments on CV and NLP tasks with various network architectures 
		(\textit{ViT-B/16}, \textit{ViT-B/32}, \textit{ViT-L/14},  \textit{ConvNeXt-Base}, and \textit{Flat-T5-Base})
		demonstrate that
		BYOM-FFT and BYOM-LoRA \textbf{outperform} existing merging methods.  
		Furthermore, 
		they achieve comparable performance to the Single-Task method but are much more parameter-efficient ($4.5\times$
		fewer parameters in reusing FFT models).
		\item BYOM-FFT
		is general and can be combined with any existing merging methods to further boost performance.
	\end{enumerate*}
	
	%%%%%%%%%%%%%%%%%%%%%%%%%%%%%%%%%
	%% Section: Related works
	%%%%%%%%%%%%%%%%%%%%%%%%%%%%%%%%%
	\section{Related Works}
	\label{sec:related-work}
	
	\textbf{Problem Formulation.}
	Consider a neural network $f(\vx; \vtheta)$ with input $\vx$ and parameters $\vtheta\in \bR^d$.
	Let $\vtheta_0$ be a pre-trained model provided on
	{torchvision}~\citep{marcel2010torchvision},  {Hugging Face},
	or {timm}~\citep{rw2019timm},
	e.g., \textit{ViT-B/32}~\citep{dosovitskiy2021an}.
	Given $T$ tasks, each task has a model finetuned from $\vtheta_0$. 
	We aim to reuse task-specific finetuned models $\{\vtheta_t: t=1,\dots, T\}$ to construct a multi-task model that can solve $T$ tasks simultaneously.
	This setting is different from multi-task learning (MTL)~\citep{kendall2018multi, ljd19, ye2021multi, ye2024first, lin2022reasonable, lin2023scale} with
	parameter-efficient tuning (e.g., LoRA~\citep{hu2021lora} and (IA)$^3$~\citep{liu2022few}).
	MTL
	needs to train the model on all task-specific data.
	Thus, it is neither data-efficient nor computation-efficient.
	Recently, many merging methods have been proposed to build a multi-task model from task-specific models.
	We first introduce three desirable properties and review existing merging methods.
	
	\begin{mybox}{Data-Efficiency (DE).}
		% \small
		An algorithm is data-efficient if it does not require any pre-training, training, or (unlabeled) testing data in merging.
	\end{mybox}
	
	\begin{mybox}{Computation-Efficiency (CE).}
		% \small
		An algorithm is computation-efficient if it does not require any re-training or gradient calculations in merging.
	\end{mybox}
	\begin{mybox}{Negligible Deterioration (ND).}
		% \small
		After merging, a multi-task model has negligible deterioration if it can reach comparable performance of the single-task method.
	\end{mybox}

	\textbf{Model Merging.}
	Task-Arithmetic~\citep{ilharco2023editing} merges all model parameters as $\vtheta^\star = \vtheta_0 + \lambda \sum_{t=1}^{T} (\vtheta_t - \vtheta_0)$, where $\lambda$ is a hyperparameter, and $\vv_t \equiv \vtheta_t - \vtheta_0$ is a task vector represents the element-wise difference between $\vtheta_t$ and $\vtheta_0$. 
	TIES-Merging~\citep{yadav2023resolving} trims low-magnitude elements in the task vector $\vv_t$ and resolves sign disagreements across task models before merging. 
	Both Task-Arithmetic and TIES-Merging
	suffer from a large deterioration in performance.
	Fisher-Merging~\citep{matena2022merging} improves uniform merging to weighted merging,
	where the weights are determined by the Fisher information matrix. 
	As the Fisher matrix requires calculating the gradient of log-likelihood loss on training data,
	Fisher-Merging is neither DE nor CE.
	RegMean~\citep{jin2023dataless} proposes to merge linear layers by solving a local linear regression problem, which requires inner product matrices of training data, thus not DE.
	
	Recently,
	AdaMerging~\citep{yang2023adamerging}
	proposes to
	learn coefficients on unlabeled testing data for merging models, while
	UncertaintyMerging~\citep{daheim2023model} proposes an uncertainty merging method based on several Fisher matrices
	computed from pre-training and training data.
	Both AdaMerging and UncertaintyMerging are neither DE nor CE.
	Moreover,
	AdaMerging and UncertaintyMerging~\citep{daheim2023model} are NOT compared with other merging methods throughout this paper because the former requires \textit{unlabeled testing data} while the latter requires \textit{pre-training data},
	which are usually unavailable in the standard merging settings~\citep{ilharco2023editing, yadav2023resolving}.
	Table~\ref{tab:compare-method} compares the proposed BYOM with existing merging methods in building multi-task models.
	As shown, only the proposed BYOM has all the desirable properties.

	%%%%%%%%%%%%%%%%%%%%%%%%%%%%%%%%%
	%% Section: Analysis
	%%%%%%%%%%%%%%%%%%%%%%%%%%%%%%%%%
	
	\section{Task Interference in Merging Models}

	As discussed in the Introduction and Figure~\ref{fig:vitb32-sparse-bubble},
	existing merging methods suffer a large performance gap compared with the Single-Task method.
	In this section,
	we introduce the task interference issue and conduct empirical analysis to show that interference cannot be resolved by simply merging without task-specific knowledge.
	
	In merging models,
	\textit{task interference} means merging one task-specific model can negatively impact the performance of other tasks.
	One consequence of task interference is the overall performance of a merged model is much worse than using multiple task-specific models.
	
	\textbf{Task interference is more serious when merging more tasks.}
	We conduct an experiment to study
	the performance of existing
	merging methods when varying the number of tasks being merged.
	For a merging method $\hA$,
	we normalize its accuracy on each task by the accuracy on the task-specific model $\vtheta_t$ and report the average normalized accuracy, i.e.,
	$\frac{1}{T}\sum_{t=1}^T \frac{\text{Accuracy on task $t$ using $\hA$} }{\text{Accuracy on task $t$ using $\vtheta_t$}}$.
	Figure \ref{fig:fft-num-task-relative-acc} 
	shows the average normalized accuracy with respect to the number of tasks being merged.
	We can see that, when merging more tasks,
	the performance of 
	existing methods (i.e., Task-Arithmetic, Fisher-Merging, RegMean, and TIES-Merging)
	decreases rapidly,
	suggesting that
	task interference is more serious.
	In contrast, only a slight performance drop is observed on the BYOM-FFT method to be proposed in Section~\ref{sec:method-fft}.
	
	\textbf{Task interference cannot be addressed by simple merging.}
	We consider merging two \textit{dissimilar} tasks \textit{DTD} and \textit{Cars} using representative merging algorithms
	(i)
	Weighting-Averaging:
	$
	\vtheta^\star = \lambda \vtheta_1 + (1-\lambda) \vtheta_2
	$,
	and (ii)
	Task-Arithmetic:
	$
	\vtheta^\star = \vtheta_0 + \lambda \sum_{t=1}^2 (\vtheta_t - \vtheta_0)
	$,
	where $\vtheta_1$ and $\vtheta_2$ denote task-specific models finetuned from \textit{ViT-B/32} on \textit{DTD} and \textit{Cars}, respectively.
	Figure \ref{fig:dis-sim-task}
	shows the average testing accuracy 
	with $\lambda\in[0,1]$.
	As can be seen, 
	for all $\lambda$'s,
	both methods always perform much worse than the Single-Task method. 
	
	One may hypothesize that task interference does not exist in merging two \textit{similar} tasks.
	To study this problem,
	we equally divide the \textit{Cars} dataset into two tasks with disjoint label sets.
	Thus, the two tasks are similar.
	% and
	% each has $98$ classes.
	Let $\vtheta_1$ and $\vtheta_2$ be the corresponding task-specific models finetuned from \textit{ViT-B/32}.
	Figure \ref{fig:sim-task}
	shows the testing accuracy of Weighting-Averaging and Task-Arithmetic
	with different $\lambda$'s.
	We can see that 
	both methods are far inferior to Single-Task.
	
	In summary, task interference always exists in merging models, even for similar tasks.
	In the next section,
	we will propose injecting task-specific knowledge into the shared model to mitigate task interference and ensure overall multi-task performance.

	%%%%%%%%%%%%%%%%%%%%%%%%%%%%%%%%%
	%% Section: Method
	%%%%%%%%%%%%%%%%%%%%%%%%%%%%%%%%%
	\section{Proposed BYOM}
	\label{sec:method}
	
	\begin{figure*}[!h]
		\centering
		\subfigure[\label{fig:vitb32_rank_bubble}\textit{ViT-B/32}.]{\includegraphics[width=0.33\textwidth]{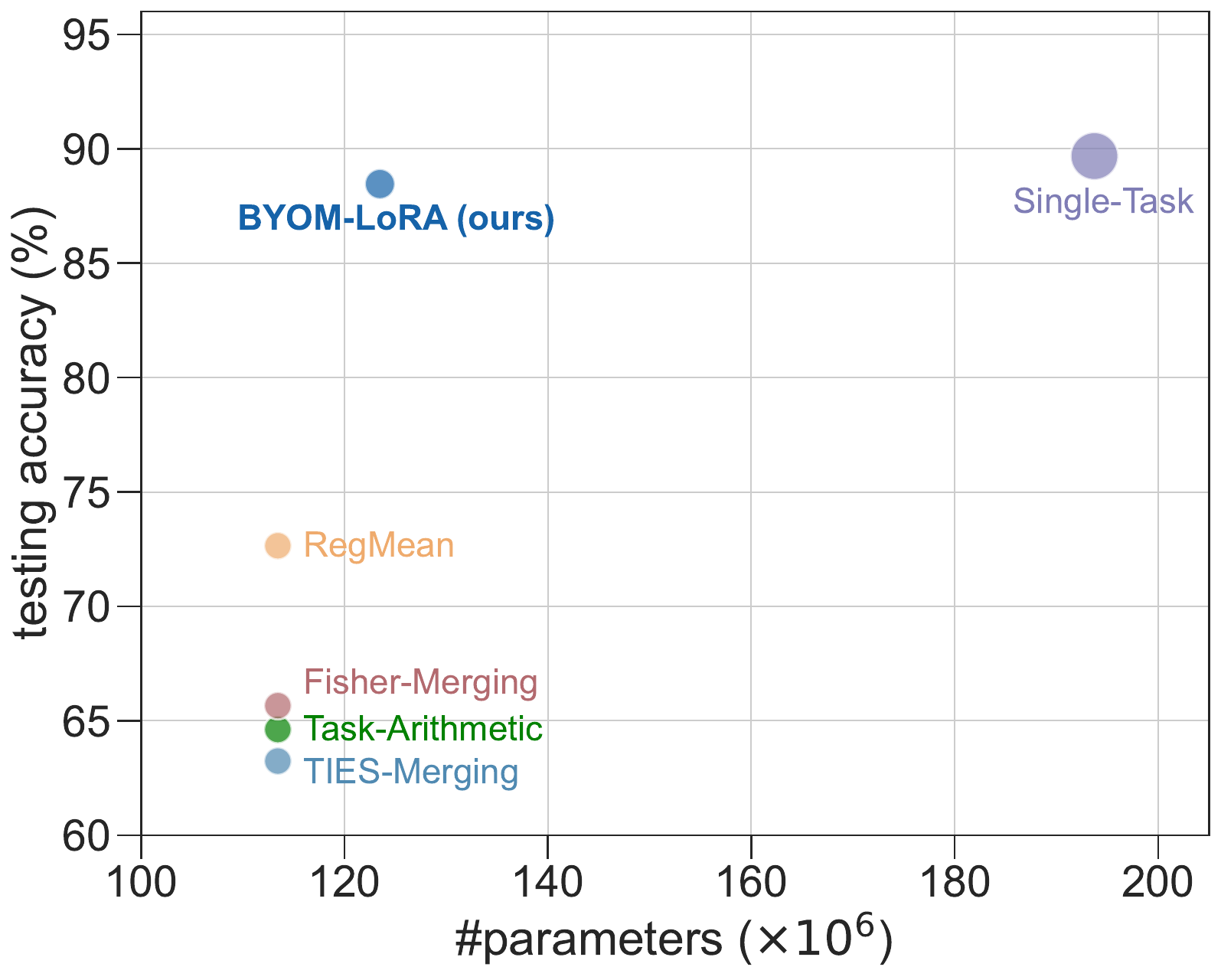}}  
		\subfigure[\label{fig:vitb16_rank_bubble}\textit{ViT-B/16}.]{\includegraphics[width=0.33\textwidth]{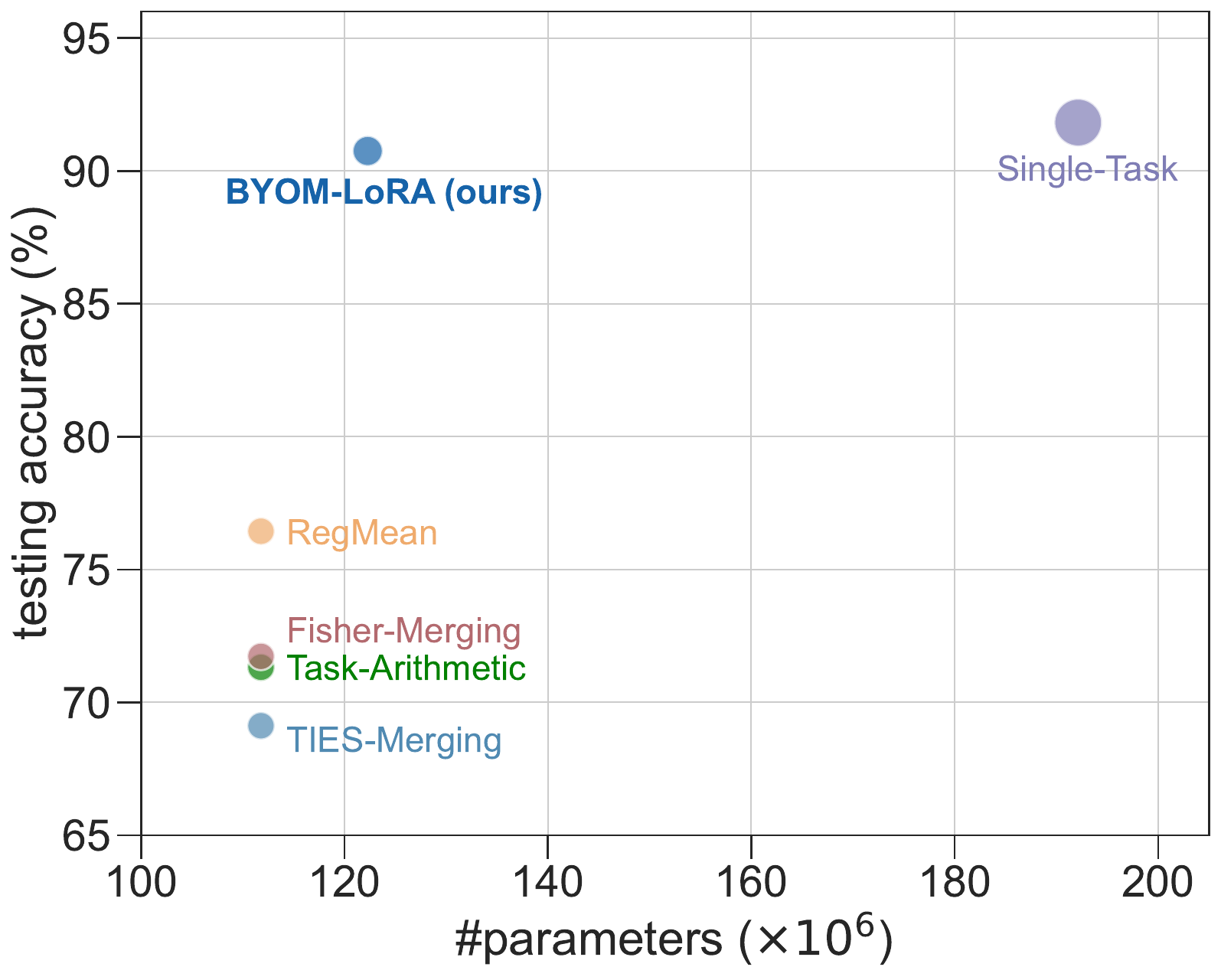}}  
		\subfigure[\label{fig:vitL14_rank_bubble}\textit{ViT-L/14}.]{\includegraphics[width=0.33\textwidth]{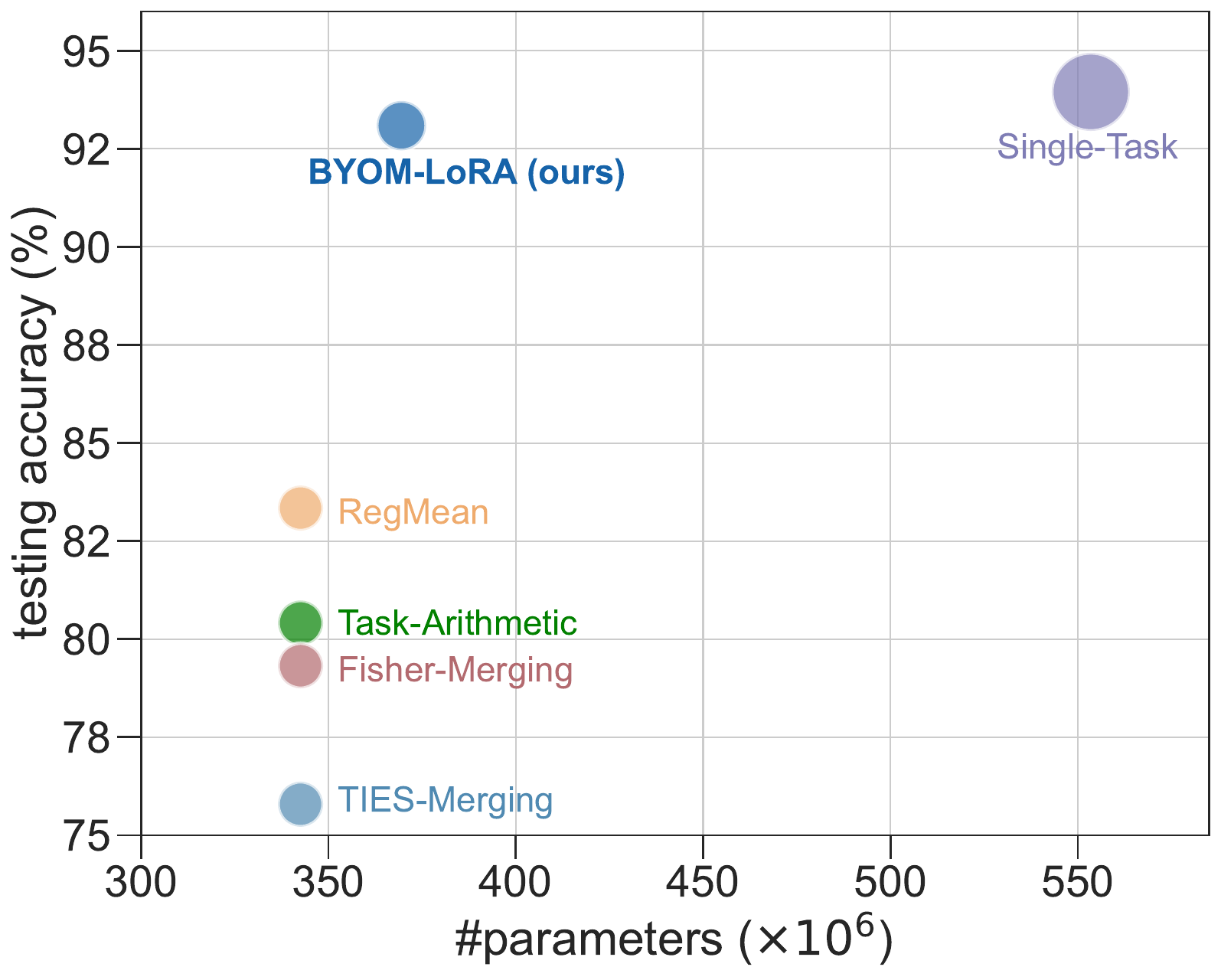}} 
		\vskip -.1in
		\caption{Testing accuracy (averaged over eight tasks) of methods reusing LoRA finetuned models.
			% (BYOM-LoRA with $q=16$). 
		}
		\label{fig:acc-rank-bubble}
		% \vskip -.1in
	\end{figure*}
	
	\subsection{Reusing Fully Finetuned Models}
	\label{sec:method-fft}
	
	For merging task-specific fully finetuned models, existing methods
	focus on merging all models into a shared model without any task-specific knowledge, leading to task interference.
	As shown in Figure \ref{fig:vitb32-sparse-bubble}, their accuracies (averaged over eight tasks) are much lower than that of Single-Task. 
	In this section, we propose BYOM-FFT to compress task-specific knowledge into a sparse vector for the shared model.
	
	We introduce
	magnitude-pruning~\citep{han2015learning, narang2016exploring, zhu2017prune} to trim 
	task-specific knowledge into a sparse vector.
	Note that 
	training-based 
	pruning methods~\citep{zhu2017prune, liu2018rethinking,wang2020pruning, zhang2022advancing, xia2022structured}
	are infeasible here
	since the training data are unavailable.
	For each task, 
	we keep the top-$m\%$ (e.g., $1\%$, $10\%$) values of the task vectors and prune the rest as
	\begin{align}
		\hat{\vv}_t(m) = \text{keep top-$m\%$ of $\vv_t$ based on magnitude}.
		\label{eq:sparse-prune}
	\end{align}
	In inference, 
	$\vtheta_0+\hat{\vv}_t(m)$
	is used as a pruned task model for the $t$th task.
	This procedure, called Post-Pruning, is
	shown in Algorithm~\ref{alg:per-sparse}.
	
	As $\vtheta_0+\hat{\vv}_t(m)$
	depends only on the $t$th task model, 
	it does not contain any shared knowledge from  the
	other tasks. 
	To address this issue,
	we propose to perform merging before pruning.
	Specifically,
	let $\vu_t = \vtheta_t - \vtheta^\star$ ($t=1, \dots, T$),
	where 
	$\vtheta^\star=\vtheta_0 + \lambda \sum_{t=1}^{T} (\vtheta_t-\vtheta_0)$
	is a merged model obtained by Task-Arithmetic. 
	We prune $\vu_t$ to $\hat{\vu}_t(m)$ 
	by keeping the top-$m\%$ values of $\vu_t$ as in Eq.~\eqref{eq:sparse-prune}.
	In inference,
	$\vtheta^\star + \hat{\vu}_t(m)$
	is used as a pruned task model,
	where $\vtheta^\star$ contains \textbf{shared knowledge} from the other tasks and $\hat{\vu}_t(m)$ contains \textbf{task-specific knowledge}. 
	The above procedure, called BYOM-FFT (\textbf{B}uilding \textbf{Y}our \textbf{O}wn \textbf{M}ulti-task models from \textbf{F}ully \textbf{F}ine\textbf{T}uned models),
	is shown in Algorithm~\ref{alg:per-sparse}.
	As pruning does not require training, 
	BYOM-FFT is \textbf{data-efficient} and \textbf{computation-efficient}.
	Moreover,
	BYOM-FFT has 
	\textbf{negligible deterioration} in performance (Figure
	\ref{fig:vitb32-sparse-bubble}).
	As
	the method for
	obtaining $\vtheta^\star$
	is flexible, any other merging algorithms (e.g.,
	Fisher-Merging,
	RegMean, and TIES-Merging) can be adopted.
	\vskip -.1in
	
	\begin{algorithm}[t]
		\caption{Post-Pruning {\color{red3}(resp. BYOM-FFT)}.}
		\label{alg:per-sparse}
		\begin{algorithmic}[1]
			\Require $m\%$;  $\vtheta_0; \vtheta_1,\dots, \vtheta_T$;
			$\lambda=0.3$;
			\State  {\color{red3} if BYOM-FFT: $\vtheta^\star = \vtheta_0 + \lambda \sum_{t=1}^{T} (\vtheta_t-\vtheta_0)$;}
			\For{$t=1, \dots, T$}
			\State $\vv_t = \vtheta_t - \vtheta_0$ {\color{red3}(resp. $\vu_t = \vtheta_t - \vtheta^\star$)};
			\State obtain  $\hat{\vv}_t(m)$ {\color{red3}(resp. $\hat{\vu}_t(m)$) } by Eq.~\eqref{eq:sparse-prune};
			\State evaluate $ \vtheta_0 + \hat{\vv}_t(m)$ {\color{red3}(resp. $\vtheta^\star + \hat{\vu}_t(m)$)} on task $t$;
			\EndFor
		\end{algorithmic}
	\end{algorithm}
	
	\subsection{Reusing LoRA Finetuned Models}
	\label{sec:lora}
	As pre-trained models are usually huge (e.g., \textit{ViT-L/14}~\citep{dosovitskiy2021an} has 343M parameters, 
	\textit{T5-base}~\citep{Colin2020t5} has 220M parameters),	
	LoRA Fine-Tuning~\citep{hu2021lora} is proposed
	to obtain task-specific models in a parameter-efficient manner.
	The finetuned task model 
	$\vtheta_ t\in \bR^{d_{\text{out}} \times d_\text{in}}$
	is decomposed as 
	\begin{align}
		\vtheta_ t = \vtheta_0 + \vA_t \vB_t^\top,
	\end{align}
	where
	$ \vA_t \in \bR^{d_{\text{out}}\times r}$, $ \vB_t \in \bR^{d_{\text{in}}\times r}$,
	and $r\ll \{d_\text{in}, d_\text{out}\}$.
	The number of parameters required in 
	LoRA fine-tuning is $r \times (d_\text{out} + d_\text{in})$, which is much smaller than
	that in
	fully fine-tuning ($d_{\text{out}} \times d_{\text{in}}$) as 
	$r$ is usually small, e.g., $r=128$.
	
	Existing methods for merging fully finetuned models can be applied directly to merging LoRA finetuned models $\{\vtheta_ t:t=1,\dots, T\}$. 
	However, they perform much worse than the Single-Task (LoRA finetuned) method (Figure \ref{fig:acc-rank-bubble}\footnote{Experimental setup is in Section~\ref{sec:expt-setup}.}). 
	Hence, using a model for all tasks without task-specific knowledge is undesirable. 
	Unlike reusing fully finetuned models, sparsifying $\vA_t \vB_t^\top$ is not parameter-efficient compared with storing $\vA_t$ and $\vB_t$ separately. 
	Hence, we use singular value decomposition (SVD) to compress task-specific LoRA matrices.
	
	To improve parameter-efficiency, we approximate $\vA_t\vB_t^\top$ by a lower-rank matrix. 
	Specifically, we first perform SVD for $\vA_t \vB_t^\top = \vU_t \vSigma_t \vV_t^\top$, where $\vU_t\in \bR^{d_\text{out}\times r}$, $\vV_t\in \bR^{d_\text{in}\times r}$, and $\vSigma_t \in \bR^{r\times r}$ is a diagonal matrix with diagonal entries sorted in descending order. Let $\vU_t(q) \in \bR^{d_\text{out}\times q}$ 
	(resp. $\vV_t(q) \in \bR^{d_\text{in}\times q}$)
	be the submatrix of the first $q$ columns of $\vU_t$
	(resp. $\vV_t$), $\vSigma_t (q) \in \bR^{q\times q}$ be the submatrix of first $q$ rows and columns of $\vSigma_t$ (corresponding to the $q$ largest singular values). The LoRA matrix $\vA_t \vB_t^{\top}$ is then approximated as $\vU_t(q) \vSigma_t (q) \vV_t(q)^\top$, where the number of parameters is reduced from $ r \times (d_\text{out} + d_\text{in})$ to $q \times (d_\text{out} + d_\text{in} + 1)$. $q$ can be much smaller than $r$ (e.g., $q=16$ compared with $r=128$, saving $8\times$ additional parameters in the LoRA matrices). 
	Moreover, BYOM-LoRA is \textbf{data-efficient} and \textbf{computation-efficient} as training is not required in the algorithm.
	In inference, $\vtheta_0 + \vU_t(q) \vSigma_t (q) \vV_t(q)^\top$ is used as the task model. 
	The procedure, called BYOM-LoRA, is shown in Algorithm \ref{alg:u-rank}.
	
	\textbf{Discussion.}
	For reusing LoRA finetuned models,
	extracting a small fraction of task-specific parameters from $\vtheta_t - \vtheta^\star$ by SVD is infeasible as $\vtheta_t - \vtheta^\star = \vtheta_0 + \vA_t \vB_t^\top - \vtheta^\star$
	is not always a rank-$r$ matrix.
	For example, 
	if $\vtheta^\star$ is obtained by Task-Arithmetic,
	$\vtheta_t - \vtheta^\star =\vA_t\vB_t^\top - \lambda\sum_{i=1}^{T} \vA_i\vB_i^\top$,
	whose rank can be $rT$.
	
	\begin{algorithm}[!h]
		\caption{BYOM-LoRA.}
		\label{alg:u-rank}
		\begin{algorithmic}[1]
			\Require
			$\vtheta_0$; LoRA matrices 
			$\{(\vA_t, \vB_t)\}_{t=1}^T$; rank $q$; 
			\For{$t=1, \dots, T$}
			\State compute $\vU_t(q),\! \vV_t(q),\! \vSigma_t(q)$ from $\vA_t\vB_t^\top\!$ by SVD;\!\!
			\State evaluate $ \vtheta_0 + \vU_t(q) \vSigma_t (q) \vV_t(q)^\top$  on task $t$;
			\EndFor
		\end{algorithmic}
	\end{algorithm}
	\vspace{-0.05in}
	
	\begin{table*}[!h]
		%	\vskip -.2in
		\centering
		\caption{Testing accuracy on eight CV tasks using \textit{ViT-B/32}.
		} 
		\label{table:vitb32}
		\resizebox{.8\textwidth}{!}{
			\renewcommand{\arraystretch}{1.15}
			\begin{NiceTabular}{c@{\hspace{.05in}}l@{\hspace{-.15in}}c@{\hspace{.08in}}c@{\hspace{.08in}}c@{\hspace{.08in}}c@{\hspace{.08in}}c@{\hspace{.08in}}c@{\hspace{.08in}}c@{\hspace{.08in}}c@{\hspace{.1in}}c@{\hspace{.1in}}c}
				\CodeBefore  
				\rectanglecolor{blue3!5}{3-1}{14-1}
				\rectanglecolor{blue3!10}{15-1}{22-1}
				\rectanglecolor{Gray}{12-2}{14-12}
				\rectanglecolor{Gray}{20-2}{22-12}
				\Body
				\arrayrulecolor{black}\specialrule{1.3pt}{.3\jot}{0.3pc}
				& & \#params (M)
				& \textit{MNI} & \textit{GTS} & \textit{SVH} & \textit{RES} & \textit{SUN} & \textit{EUR} & \textit{DTD} & \textit{CAR} & Avg \\
				\arrayrulecolor{black}\specialrule{1.3pt}{.3\jot}{0.3pc}
				%%%%
				%%%% fll ft
				%%%%
				&Pre-Trained & 113 & 48.25 & 32.56 & 31.61 & 60.65 & 63.18 & 45.11 & 43.99 & 59.74 & 48.14 \\
				\arrayrulecolor{black}\specialrule{1.3pt}{.3\jot}{0.3pc}
				\multirow{13}{*}{\STAB{\rotatebox[origin=c]{90}{Fully FT}}}
				&{Single-Task}  & {908} &{ 99.72} & {99.23} & { 97.42} & {95.56} & {75.03} & {99.00} & {79.47} & {78.73} & {90.52} \\ 
				& {\color{rev} MTL} & {\color{rev}113} & {\color{rev}99.45} & {\color{rev}98.91} & {\color{rev}95.80} & {\color{rev}93.90} & {\color{rev}72.85} & {\color{rev}98.22} &{ \color{rev}77.87} & {\color{rev}74.44} & {\color{rev}88.93} \\
				\cmidrule{2-12}
				&Task-Arithmetic & 113 & 93.27 & 65.99 & 71.62 & 71.57 & 63.63 & 78.41 & 51.76 & 61.50 & 69.72 \\
				&Fisher-Merging & 113 & 80.71 & 75.15 & 74.08 & 70.24 & 65.25 & 81.48 & 49.84 & 62.90 & 69.96 \\
				&RegMean & 113 & 92.55 & 65.12 & 75.48 & 75.56 & 65.72 & 84.33 & 56.01 & 64.54 & 72.41 \\
				&TIES-Merging & 113 & 97.79 & 75.30 & 84.10 & 70.71 & 59.24 & 75.89 & 53.51 & 58.72 & 71.91 \\
				\cmidrule{2-12}
				&	Post-Pruning ($1\%$) & 123 & 58.41 & 40.61 & 39.38 & 67.08 & 66.63 & 56.26 & 48.83 & 63.95 & 55.14 \\
				&	Post-Pruning ($5\%$) & 159 & 95.82 & 78.61 & 74.35 & 83.67 & 71.60 & 85.81 & 62.39 & 72.73 & 78.12 \\
				&Post-Pruning ($10\%$) & 204 & 99.17 & 95.30 & 93.85 & 92.13 & \textbf{74.39 }& 96.37 & 71.97 & \textbf{77.09} & 87.53 \\
				\cmidrule{2-12}
				&BYOM-FFT ($1\%$) & 123 & 96.17 & 76.33 & 79.27 & 78.03 & 66.88 & 84.89 & 58.03 & 65.99 & 75.70 \\
				&	BYOM-FFT ($5\%$) & 159 & 99.12 & 92.66 & 91.86 & 88.48 & 71.35 & 94.85 & 67.77 & 73.08 & 84.90 \\
				&BYOM-FFT ($10\%$) & 204 & \textbf{99.49} & \textbf{97.57} & \textbf{95.92} & \textbf{93.00} & 73.52 & \textbf{97.63} & \textbf{72.98} & 76.92 & \textbf{88.38} \\
				\arrayrulecolor{black}\specialrule{1.3pt}{.3\jot}{0.3pc}
				%%%%
				%%%% lora ft
				%%%%
				\multirow{8}{*}{\STAB{\rotatebox[origin=c]{90}{LoRA FT}}}
				&Single-Task  & 194 & 99.61 & 98.71 & 97.34 & 95.57 & 73.42 & 98.63 & 76.91 & 77.25 & 89.68 \\
				\cmidrule{2-12}
				&Task-Arithmetic & 113 & 86.90 & 51.44 & 66.50 & 68.16 & 62.32 & 76.19 & 48.62 & 56.85 & 64.62 \\
				&Fisher-Merging & 113 & 86.71 & 53.85 & 62.44 & 71.19 & 65.16 & 72.67 & 50.37 & 62.88 & 65.66 \\
				&RegMean & 113 & 94.45 & 60.10 & 81.11 & 74.57 & 65.10 & 88.15 & 53.72 & 63.97 & 72.65 \\
				&TIES-Merging & 113 & 82.48 & 45.89 & 58.95 & 70.67 & 65.20 & 71.11 & 49.15 & 62.44 & 63.24 \\
				\cmidrule{2-12}
				%				&BYOM-LoRA (2) & 115 & 92.55 & 85.91 & 90.16 & 81.51 & 67.58 & 87.63 & 59.47 & 60.54 & 78.17 \\
				&BYOM-LoRA (4) & 116 & 99.16 & 92.04 & 93.98 & 86.48 & 68.61 & 95.37 & 65.37 & 62.74 & 82.97 \\
				&BYOM-LoRA (8) & 118 & 99.54 & 96.23 & 96.45 & 92.16 & 70.33 & 98.26 & 72.55 & 67.35 & 86.61 \\
				&BYOM-LoRA (16) & 123 & \textbf{99.62} & \textbf{97.99} & \textbf{97.08} & \textbf{94.56} & \textbf{72.29} &\textbf{ 98.37} & \textbf{76.44} &\textbf{ 71.31} & \textbf{88.46} \\
				\arrayrulecolor{black}\specialrule{1.3pt}{.3\jot}{0.3pc}
		\end{NiceTabular}}
	\end{table*}

	%%%%%%%%%%%%%%%%%%%%%%%%%%%%%%%%%
	%% Section: Experiments
	%%%%%%%%%%%%%%%%%%%%%%%%%%%%%%%%%
	
	\section{Experiments}
	\subsection{Evaluation on CV Tasks} \label{sec:expt-setup}
	
	\textbf{Datasets and models.}
	Following~\citep{ilharco2023editing, yadav2023resolving},
	experiments are conducted on 
	eight image classification tasks:
	\textit{MNIST} (denoted by MNI)~\citep{lecun2010mnist},
	\textit{GTSRB} (denoted by GTS)~\citep{stallkamp2011german}, 
	\textit{SVHN} (denoted by SVH)~\citep{Netzer2011},
	\textit{RESISC45} (denoted by RES)~\citep{cheng2017remote},
	\textit{SUN397} (denoted by SUN)~\citep{xiao2016sun},
	\textit{EuroSAT} (denoted by EUR)~\citep{helber2019eurosat},
	\textit{DTD}~\citep{Cimpoi2014},
	and \textit{Cars} (denoted by CAR)~\citep{Krause2013}.
	
	Following \citep{ilharco2023editing},
	we adopt three variants of the CLIP model~\citep{radford2021learning}
	with \textit{ViT} models~\citep{dosovitskiy2021an} including \textit{ViT-B/32}, \textit{ViT-B/16}, and \textit{ViT-L/14}
	as image encoders.
	
	\textbf{Baselines.}
	We compare with 
	\begin{enumerate*}[(i), series = tobecont, itemjoin = \quad]
		\item Pre-Trained Model $\vtheta_0$;
		\item multiple Single-Task fully/LoRA finetuned models (Single-Task);
		\item {\color{rev}Multi-Task Learning (MTL) \citep{zhang2021survey} which requires all task data for training a model};
		and the state-of-the-art merging methods including
		\item Task-Arithmetic~\citep{ilharco2023editing} 
		merges model parameters by uniform averaging;
		\item Fisher-Merging~\citep{matena2022merging}
		takes weighted averaging based on Fisher information matrix;
		\item RegMean~\citep{jin2023dataless}
		merges linear layers by solving a local linear regression problem;
		\item TIES-Merging~\citep{yadav2023resolving}
		trims the task vectors and resolves the  sign disagreements
		before aggregating parameters.
		Note that
		AdaMerging~\citep{yang2023adamerging} and UncertaintyMerging~\citep{daheim2023model} are NOT compared because the former needs \textit{unlabeled testing data} while the latter requires \textit{pre-training data},
		which are unavailable in the standard merging setting~\citep{ilharco2023editing, yadav2023resolving}.
	\end{enumerate*}
	
	\textbf{Results.} 
	Tables~\ref{table:vitb32}, \ref{table:vitb16} (in Appendix \ref{sec:expt-vit-b16}), and \ref{table:vitL14} show the testing accuracies
	on eight data sets using \textit{ViT-B/32},
	\textit{ViT-B/16},
	and \textit{ViT-L/14},
	respectively.
	As can be seen,
	for merging fully finetuned models,
	BYOM-FFT with $m\%=1\%$ (denoted BYOM-FFT(1\%))
	performs better than existing merging methods,
	demonstrating that injecting task-specific knowledge into the shared model is useful.
	Moreover,
	by keeping top-$10\%$ values,
	BYOM-FFT
	has negligible deterioration in performance compared to 
	Single-Task  
	but is more parameter-efficient ($4.5\times$ fewer parameters).
	Compared with 
	Post-Pruning,
	BYOM-FFT
	achieves higher average accuracy over eight tasks,
	showing that
	the shared knowledge is beneficial.
	
	\begin{table*}[!h]
		\centering
		\caption{Testing accuracy on eight CV tasks  using \textit{ViT-L/14}.
		} 
		\label{table:vitL14}
		\resizebox{.8\textwidth}{!}{
			\renewcommand{\arraystretch}{1.15}
			\begin{NiceTabular}{c@{\hspace{.05in}}l@{\hspace{-.15in}}c@{\hspace{.08in}}c@{\hspace{.08in}}c@{\hspace{.08in}}c@{\hspace{.08in}}c@{\hspace{.08in}}c@{\hspace{.08in}}c@{\hspace{.08in}}c@{\hspace{.1in}}c@{\hspace{.1in}}c}
				\CodeBefore  
				\rectanglecolor{blue3!5}{3-1}{14-1}
				\rectanglecolor{blue3!10}{15-1}{22-1}
				\rectanglecolor{Gray}{12-2}{14-12}
				\rectanglecolor{Gray}{20-2}{22-12}
				\Body
				\arrayrulecolor{black}\specialrule{1.3pt}{.3\jot}{0.3pc}
				& & \#params (M)
				& \textit{MNI} & \textit{GTS} & \textit{SVH} & \textit{RES} & \textit{SUN} & \textit{EUR} & \textit{DTD} & \textit{CAR} & Avg \\
				\arrayrulecolor{black}\specialrule{1.3pt}{.3\jot}{0.3pc}
				%%%%
				%%%% fll ft
				%%%%
				& Pre-Trained & 343 & 76.36 & 50.55 & 58.45 & 71.05 & 68.28 & 62.41 & 55.32 & 77.73 & 65.02 \\
				\arrayrulecolor{black}\specialrule{1.3pt}{.3\jot}{0.3pc}
				\multirow{13}{*}{\STAB{\rotatebox[origin=c]{90}{Fully FT}}}
				&Single-Task & 2,740 & 99.77 & 99.33 & 98.12 & 97.30 & 82.13 & 99.26 & 84.68 & 92.36 & 94.12 \\
				& {\color{rev}MTL} & {\color{rev}343}& {\color{rev}99.63} & {\color{rev}99.07} & {\color{rev}97.57} & {\color{rev}96.32} & {\color{rev}80.84} & {\color{rev}99.19} & {\color{rev}84.36} & {\color{rev}90.64} & {\color{rev}93.45}\\
				\cmidrule{2-12}
				&Task-Arithmetic & 343 & 98.95 & 85.80 & 87.20 & 86.60 & 73.84 & 94.48 & 65.69 & 83.68 & 84.53 \\
				&Fisher-Merging & 343 & 96.98 & 69.43 & 78.20 & 82.33 & 72.18 & 91.04 & 62.07 & 82.43 & 79.33 \\
				&RegMean & 343 & 98.42 & 81.37 & 88.03 & 85.27 & 72.77 & 95.37 & 65.74 & 84.09 & 83.88 \\
				& TIES-Merging & 343 & 99.01 & 81.34 & 89.42 & 89.49 & 76.18 & 95.96 & 68.24 & 86.83 & 85.81 \\
				\cmidrule{2-12}
				&Post-Pruning ($1\%$) & 370 & 88.11 & 57.55 & 67.26 & 78.27 & 71.40 & 75.78 & 59.89 & 82.04 & 72.54 \\
				&Post-Pruning ($5\%$) & 480 & 99.07 & 84.66 & 87.85 & 92.75 & 77.40 & 97.48 & 72.02 & 88.96 & 87.52 \\
				&Post-Pruning ($10\%$) & 617 & 99.67 & 96.95 & 96.86 & 96.25 & 80.56 & \textbf{99.04} & 79.31 & \textbf{91.54} & 92.52 \\
				\cmidrule{2-12}
				&BYOM-FFT ($1\%$) & 370 & 99.17 & 90.67 & 90.99 & 89.62 & 75.55 & 96.30 & 69.36 & 86.06 & 87.21 \\
				&BYOM-FFT ($5\%$) & 480 & 99.62 & 96.46 & 95.87 & 94.41 & 78.90 & 98.41 & 76.76 & 89.14 & 91.20 \\
				&BYOM-FFT ($10\%$) & 617 & \textbf{99.74} & \textbf{98.43} & \textbf{97.43} &\textbf{ 96.37} & \textbf{80.79} & 98.93 & \textbf{80.53} & 90.72 & \textbf{92.87} \\
				\arrayrulecolor{black}\specialrule{1.3pt}{.3\jot}{0.3pc}
				%%%%
				%%%% lora ft
				%%%%
				\multirow{8}{*}{\STAB{\rotatebox[origin=c]{90}{LoRA FT}}}
				&	Single-Task & 553 & 99.78 & 99.28 & 98.02 & 97.13 & 81.79 & 99.04 & 84.52 & 92.08 & 93.95 \\
				\cmidrule{2-12}
				& Task-Arithmetic & 343 & 97.59 & 72.35 & 81.47 & 83.03 & 72.40 & 91.59 & 62.45 & 82.42 & 80.41 \\
				&Fisher-Merging & 343 & 96.98 & 69.40 & 78.18 & 82.32 & 72.18 & 91.00 & 62.07 & 82.43 & 79.32 \\
				&RegMean & 343 & 98.53 & 80.39 & 84.83 & 85.70 & 72.90 & 95.41 & 65.05 & 83.93 & 83.34 \\
				&TIES-Merging & 343 & 94.72 & 61.36 & 74.20 & 79.43 & 71.22 & 84.00 & 60.05 & 81.36 & 75.79 \\
				\cmidrule{2-12}
				%				&BYOM-LoRA (2) & 346 & 99.20 & 94.13 & 95.01 & 89.89 & 75.26 & 96.85 & 72.34 & 81.23 & 87.99 \\
				&BYOM-LoRA (4) &  349 & 99.53 & 97.47 & 96.98 & 93.32 & 76.61 & 98.63 & 76.33 & 84.07 & 90.37 \\
				&BYOM-LoRA (8) & 356 & 99.76 & 98.48 & 97.80 & 95.75 & 78.23 & 98.81 & 80.85 & 87.53 & 92.15 \\
				& BYOM-LoRA (16) & 369 & \textbf{99.78} & \textbf{98.92} & \textbf{98.02} & \textbf{96.56} & \textbf{79.91} & \textbf{99.04 }&\textbf{ 82.93} & \textbf{89.55} & \textbf{93.09} \\
				\arrayrulecolor{black}\specialrule{1.3pt}{.3\jot}{0.3pc}
		\end{NiceTabular}}
		\vskip -.2in
	\end{table*}
	
	\begin{figure*}[!h]
		\vskip -.2in
		\centering
		%		\vskip -.2in
		\!\!
		\subfigure[\label{fig:tsne_tv}TaskArithmetic.]{\includegraphics[width=0.16\textwidth]{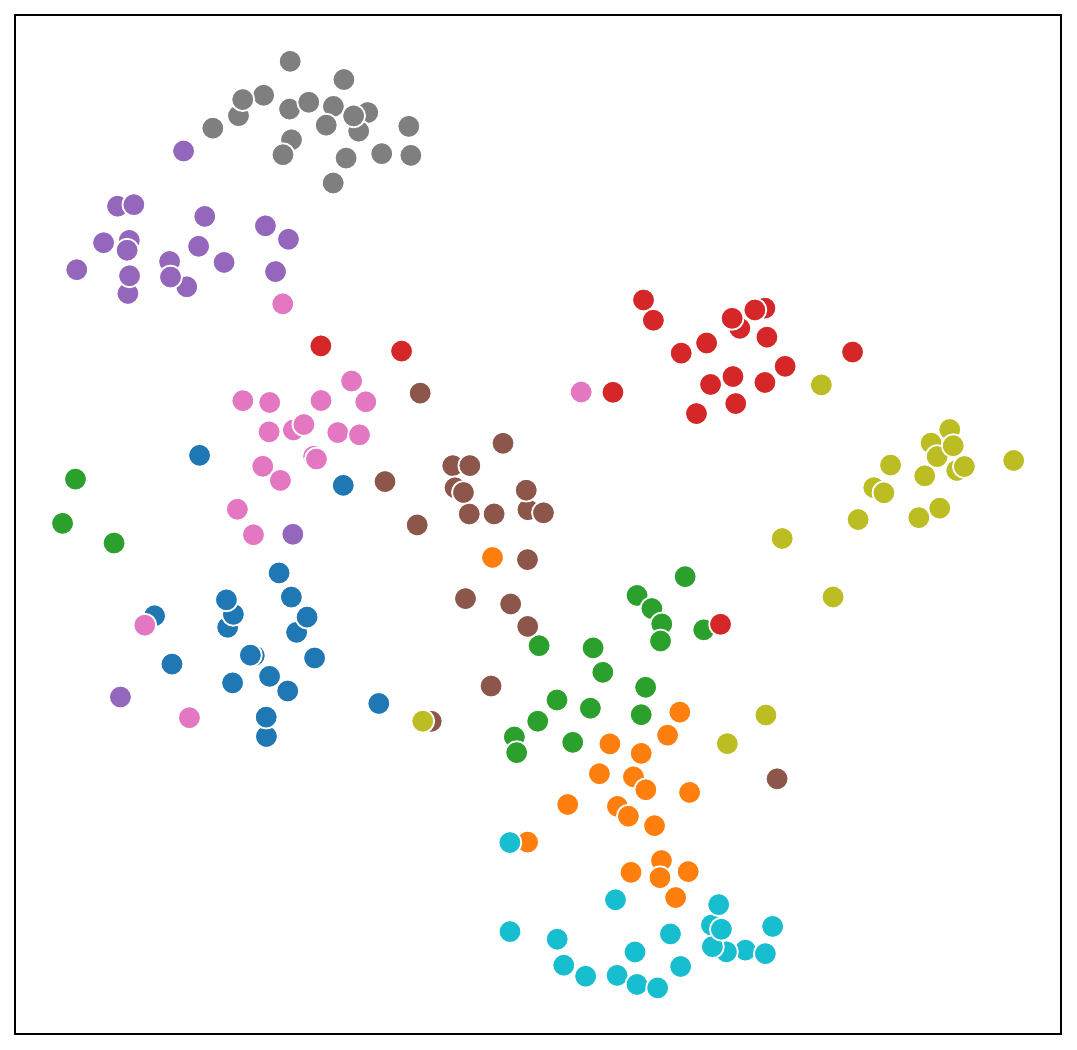}} \!\!
		\subfigure[\label{fig:tsne_fm}FisherMerging.]{\includegraphics[width=0.16\textwidth]{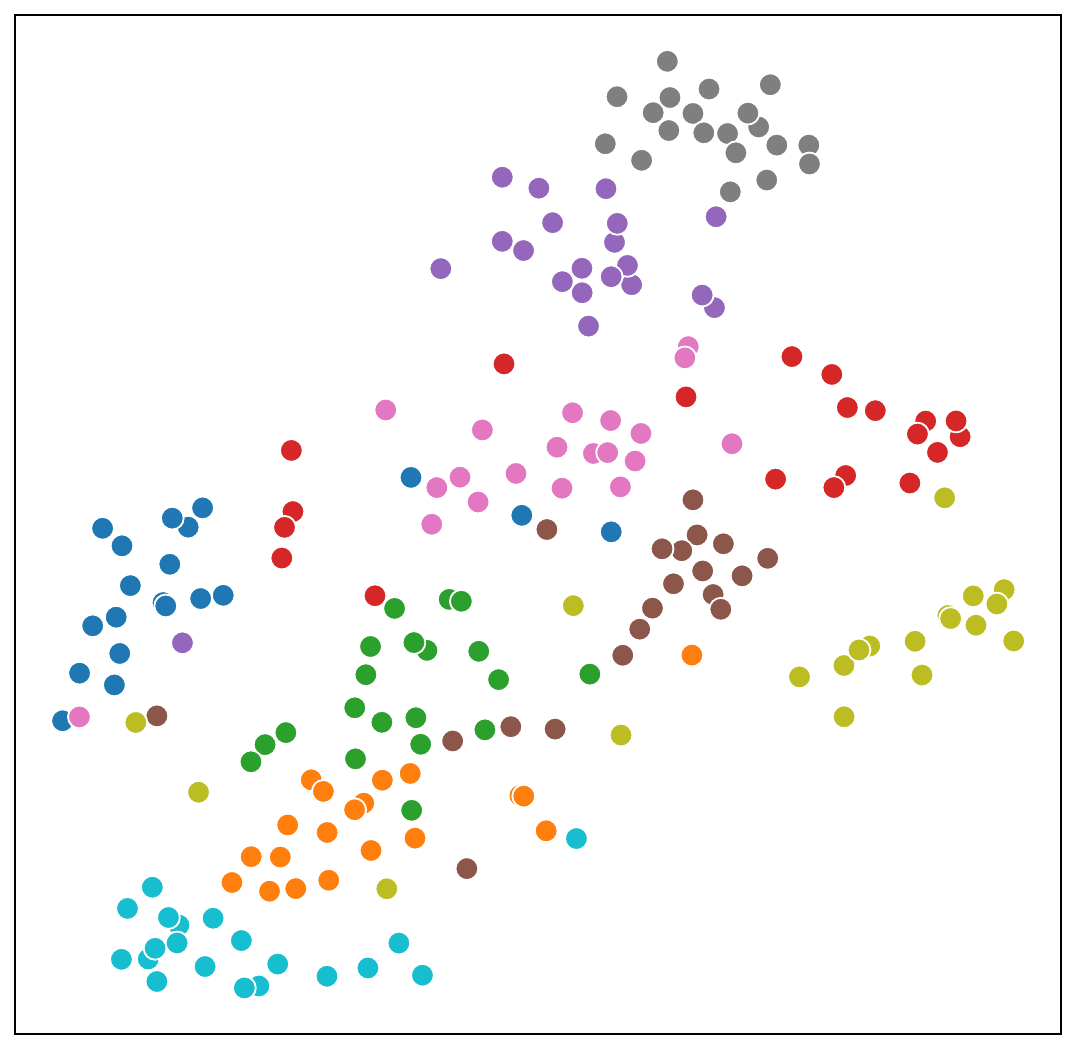}} \!\!
		\subfigure[\label{fig:tsne_rm}RegMean.]{\includegraphics[width=0.16\textwidth]{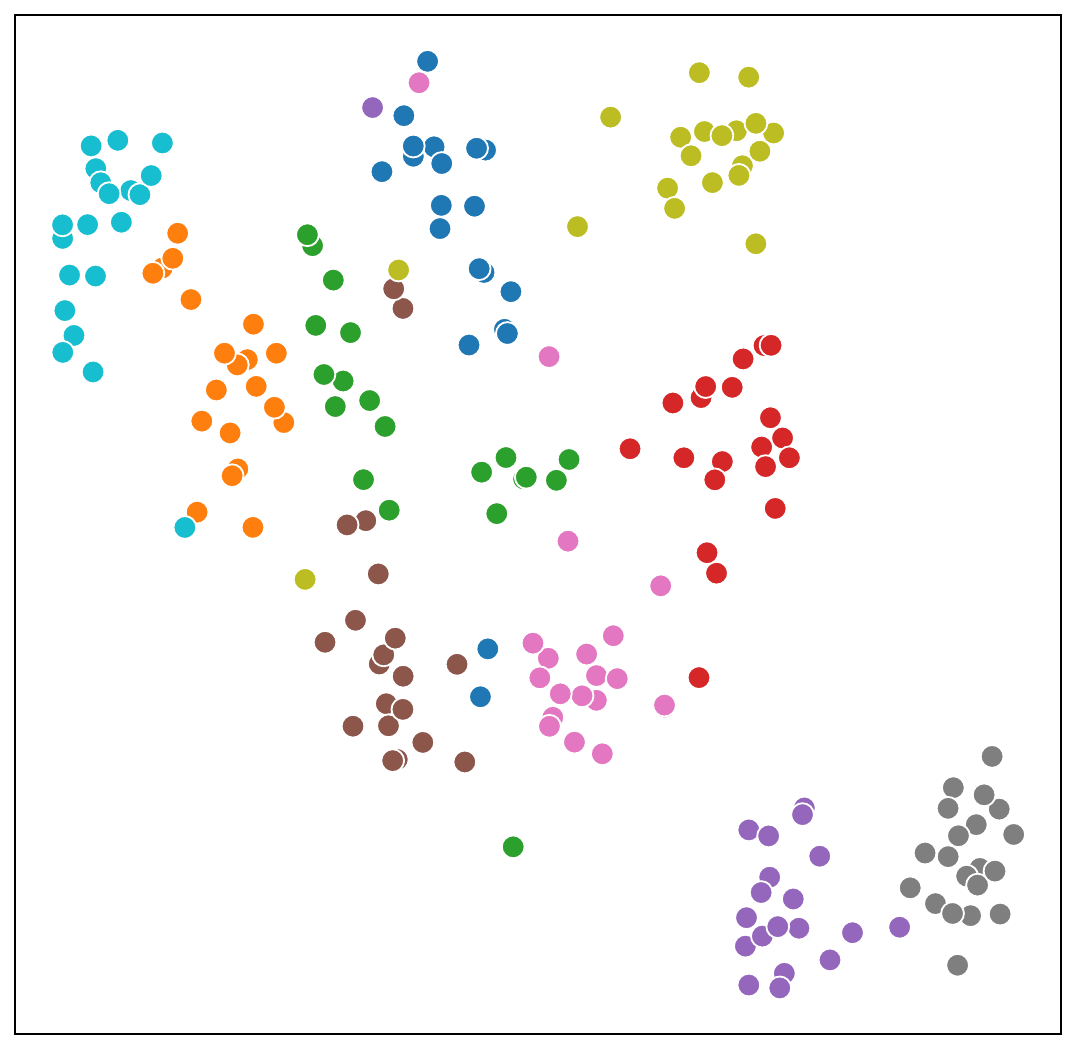}} \!\! 
		\subfigure[\label{fig:tm}TiesMerging.]{\includegraphics[width=0.16\textwidth]{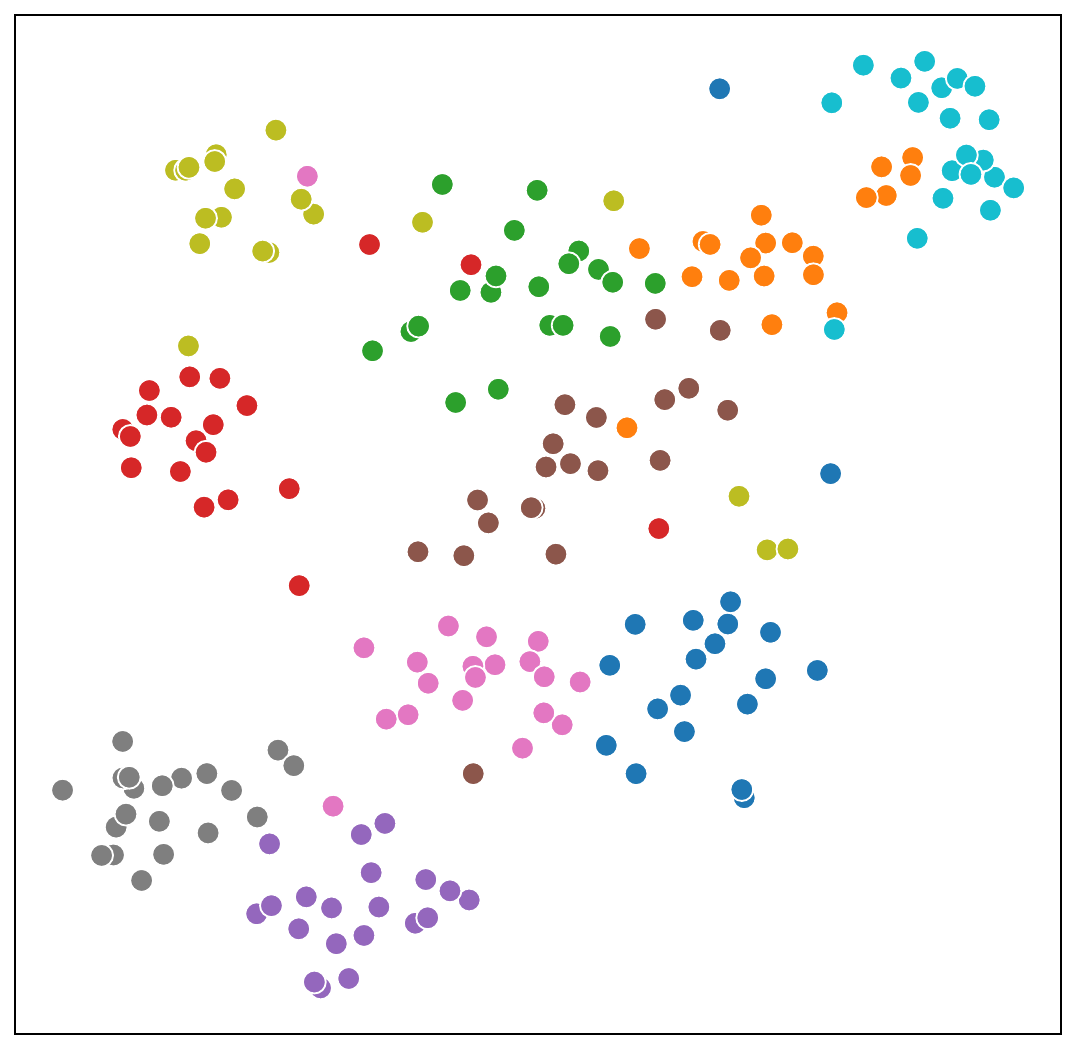}} \!\! 
		\subfigure[\label{fig:sparse_init}Post-Pruning.]{\includegraphics[width=0.16\textwidth]{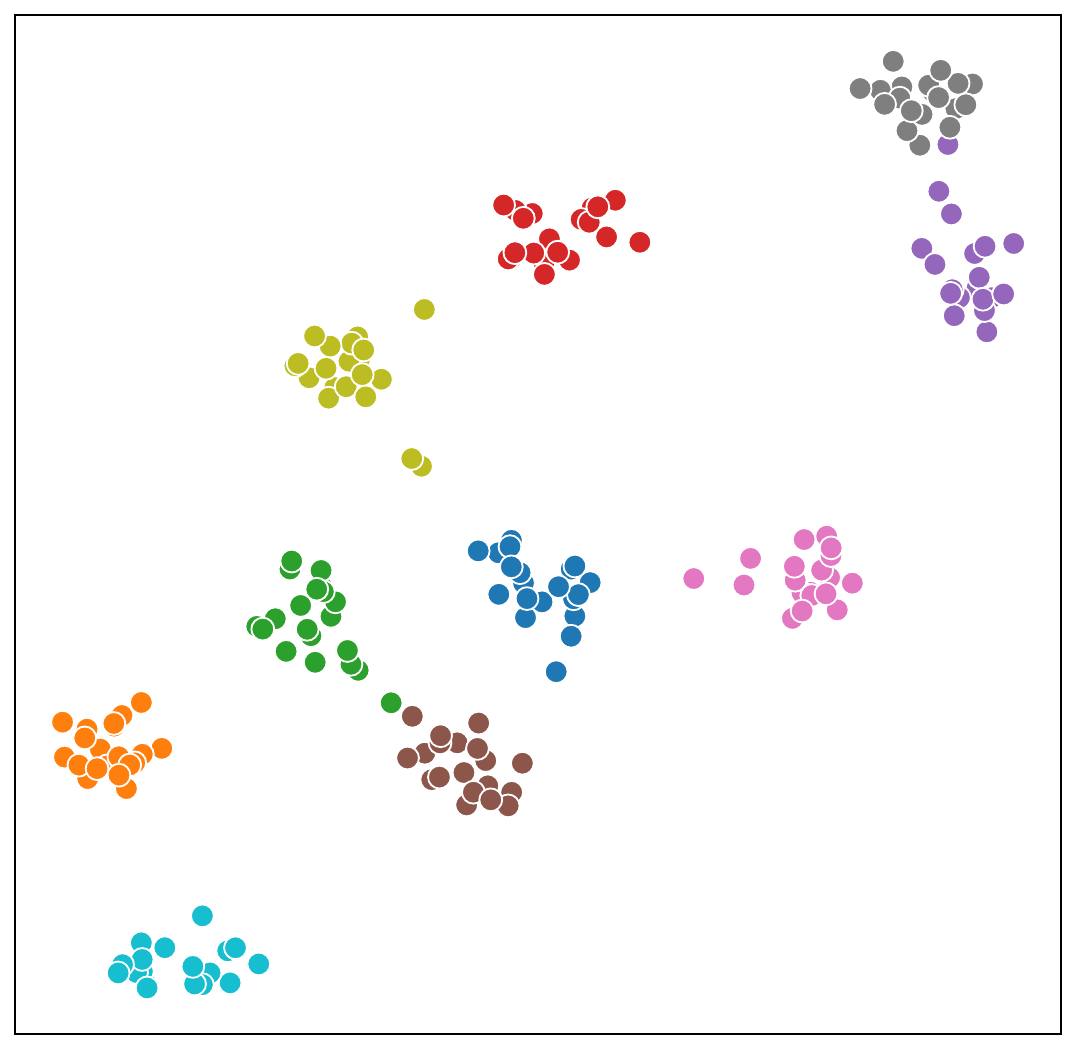}} \!\!
		\subfigure[\label{fig:spase_star}BYOM-FFT.]{\includegraphics[width=0.16\textwidth]{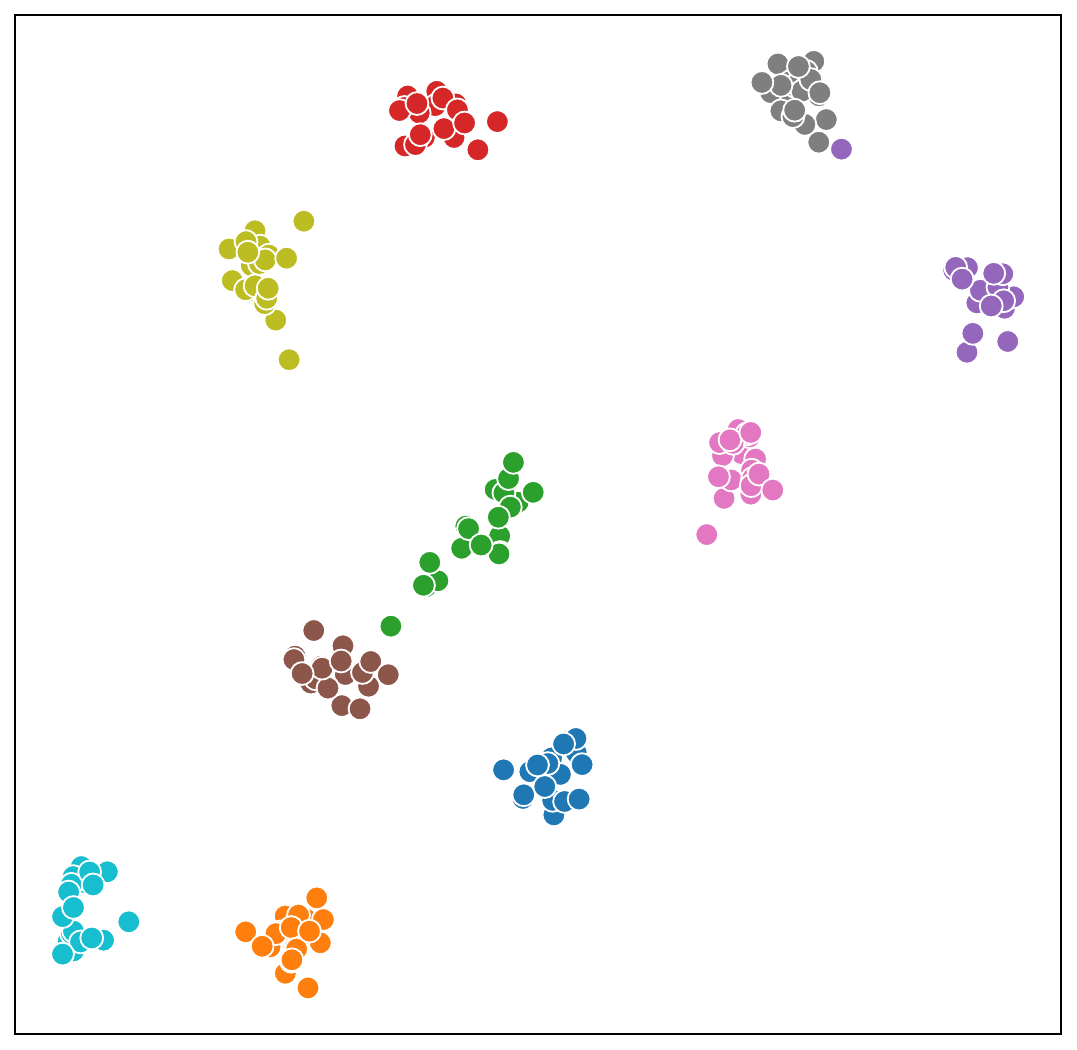}} \!\!
		\vskip -.15in
		\caption{t-SNE of samples from \textit{EuroSAT} for methods reusing fully finetuned \textit{ViT-B/32} models.}
		\label{fig:fft_tsne}
		\vskip -.15in
	\end{figure*}
	
	As for merging LoRA finetuned models, we can see that 
	BYOM-LoRA with $q=4$ (denoted BYOM-LoRA(4)) outperforms existing merging methods by a large margin,
	demonstrating the usefulness of task-specific knowledge. 
	Furthermore, 
	BYOM-LoRA(16) achieves comparable performance with Single-Task, but is more parameter-efficient ($1.6\times$ fewer parameters). 
	Compared with BYOM-FFT(10\%) and Post-Pruning(10\%), 
	BYOM-LoRA (16) performs better but has $1.7\times$ fewer parameters. 
	Moreover, BYOM-LoRA(16) achieves comparable performance with Single-Task (Fully FT) but has $7.4\times$ fewer parameters, showing that reusing the LoRA finetuned models is better. 
	Furthermore, compared with the Pre-Trained model, BYOM-LoRA(16) with only 10M more parameters almost doubles the accuracy for the \textit{ViT-B/32} model. 
	
	Figure~\ref{fig:fft_tsne} visualizes the t-SNE~\citep{van2008visualizing} embeddings extracted from 200 images (20 images per class) randomly sampled from \textit{EuroSAT} for methods reusing fully finetuned \textit{ViT-B/32} models. 
	As can be seen, BYOM-FFT($10\%$) has more compact and separable structures than existing merging methods, demonstrating that BYOM-FFT extracts more discriminative features. 
	Furthermore, clusters of BYOM-FFT are denser than Post-Pruning.
	Figure~\ref{fig:lora_tsne} in Appendix \ref{sec:vis-lora} shows the t-SNE of embeddings for methods reusing LoRA finetuned \textit{ViT-B/32} models. 
	As can be seen, BYOM-LoRA(16) has a more compact and separable structure than existing merging methods.

	\subsection{Evaluation on NLP Tasks} \label{sec:exp_nlp}
	
	We conduct experiments on four standard text classification data sets:
	\textit{MRPC}~\citep{dolan2004unsupervised}, \textit{RTE}~\citep{wang2018glue}, \textit{SST-2}~\citep{socher2013recursive},
	and \textit{QNLI}~\citep{wang2018glue}.
	We adopt \textit{Flan-T5-base} \citep{chung2022scaling}
	as the model
	for text classification.
	
	Table~\ref{table:nlp-vitb16} shows the testing accuracy. 
	As can be seen, for merging fully finetuned models, 
	by keeping top-$10\%$ values, 
	BYOM-FFT largely outperforms existing merging methods,
	showing that introducing task-specific knowledge is beneficial. 
	Furthermore,
	BYOM-FFT achieves negligible deterioration in performance compared to Single-Task but is much more parameter-efficient ($2.8\times$ few parameters). 
	Compared with Post-Pruning, BYOM-FFT is better, demonstrating the usefulness of the shared knowledge. 
	As for reusing LoRA finetuned models, BYOM-LoRA outperforms existing merging methods.
	Moreover,
	BYOM-LoRA with $q=8$ or $16$ achieves almost the same performance as Single-Task (LoRA FT) but has fewer parameters. 
	
	\begin{table}[!h]
		\vskip -.15in
		\centering
		\caption{Testing accuracy on four NLP tasks using \textit{Flan-T5-base}.
		} 
		\label{table:nlp-vitb16}
		\resizebox{.49\textwidth}{!}{
			\renewcommand{\arraystretch}{1.15}
			\begin{NiceTabular}{cl@{\hspace{-.05in}}c@{\hspace{.05in}}ccccc}
				\CodeBefore  
				\rectanglecolor{blue3!5}{3-1}{13-1}
				\rectanglecolor{blue3!10}{14-1}{21-1}
				\rectanglecolor{Gray}{11-2}{13-12}
				\rectanglecolor{Gray}{19-2}{21-12}
				\Body
				\arrayrulecolor{black}\specialrule{1.3pt}{.3\jot}{0.3pc}
				&&\#params(M)
				& \textit{MRPC} & \textit{RTE} & \textit{SST-2} & \textit{QNLI} & Avg\\
				\arrayrulecolor{black}\specialrule{1.3pt}{.3\jot}{0.3pc}
				%%%%
				%%%% fll ft
				%%%%
				& Pre-Trained & 225 & 75.33 & 57.04 & 52.64 & 66.59 & 62.90 \\
				\arrayrulecolor{black}\specialrule{1.3pt}{.3\jot}{0.3pc}
				\multirow{11}{*}{\STAB{\rotatebox[origin=c]{90}{Fully FT}}}
				&Single-Task & 894 & 89.30 & 79.06 & 94.72 & 93.00 & 89.02 \\
				\cmidrule{2-8}
				&Task-Arithmetic & 225 & 82.29 & 73.29 & 93.23 & 88.16 & 84.24 \\
				&Fisher-Merging & 225 & 80.61 & 70.04 & 92.66 & 85.63 & 82.23 \\
				&RegMean & 225 & 84.52 & 76.53 & 92.55 & 91.16 & 86.19 \\
				&TIES-Merging  & 225 & 86.70 & 74.73 & 93.23 & 84.13 & 84.70\\
				\cmidrule{2-8}
				&Post-Pruning ($1\%$) & 234 & 75.52 & 62.45 & 69.72 & 81.90 & 72.40 \\
				&Post-Pruning ($5\%$) & 270 & 81.23 & 68.23 & 92.66 & 90.28 & 83.10 \\
				&Post-Pruning ($10\%$) & 314 & 86.26 & 77.62 & 94.04 & 91.69 & 87.40 \\
				\cmidrule{2-8}
				&BYOM-FFT ($1\%$) & 234 & 83.62 & 75.81 & 93.81 & 89.86 & 85.77 \\
				&BYOM-FFT ($5\%$) & 270 & 86.63 & 78.34 & 94.04 & 91.43 & 87.61 \\
				&BYOM-FFT ($10\%$) & 314 & \textbf{87.58} & \textbf{78.70 }& \textbf{94.27 }& \textbf{91.84} & \textbf{88.10} \\
				\arrayrulecolor{black}\specialrule{1.3pt}{.3\jot}{0.3pc}
				%%%%
				%%%% lora ft
				%%%%
				\multirow{8}{*}{\STAB{\rotatebox[origin=c]{90}{LoRA FT}}}
				
				& Single-Task & 281 & 87.87 & 79.42 & 93.92 & 92.13 & 88.34 \\
				\cmidrule{2-8}
				& Task-Arithmetic & 225 & 82.87 & 74.01 & 92.55 & 86.89 & 84.08\\
				&Fisher-Merging & 225 &  80.01 & 76.17 & 91.86 & 85.17 & 83.30\\
				& RegMean & 225 & 81.09 & 75.81 & 92.32 & 91.41 & 85.16 \\
				& TIES-Merging  & 225 & 84.29 & 72.56 & 92.55 & 81.11 & 82.63\\
				\cmidrule{2-8}
				&BYOM-LoRA (4) & 227 & 86.26 & 78.70 & 93.69 & 91.96 & 87.65\\
				&BYOM-LoRA (8) & 229 & 86.80 & 79.06 & 93.81 & 92.04 & 87.93\\
				&BYOM-LoRA (16) & 232 & 87.24 & 79.42 & 93.81 & 92.07 & 88.14\\
				\arrayrulecolor{black}\specialrule{1.3pt}{.3\jot}{0.3pc}
			\end{NiceTabular}
		}
		\vskip -.2in
	\end{table}
	
	\subsection{Usefulness of Integrating BYOM-FFT into Existing Merging Methods}
	
	\begin{figure}[!h]
		\centering
		\includegraphics[width=0.45\textwidth]{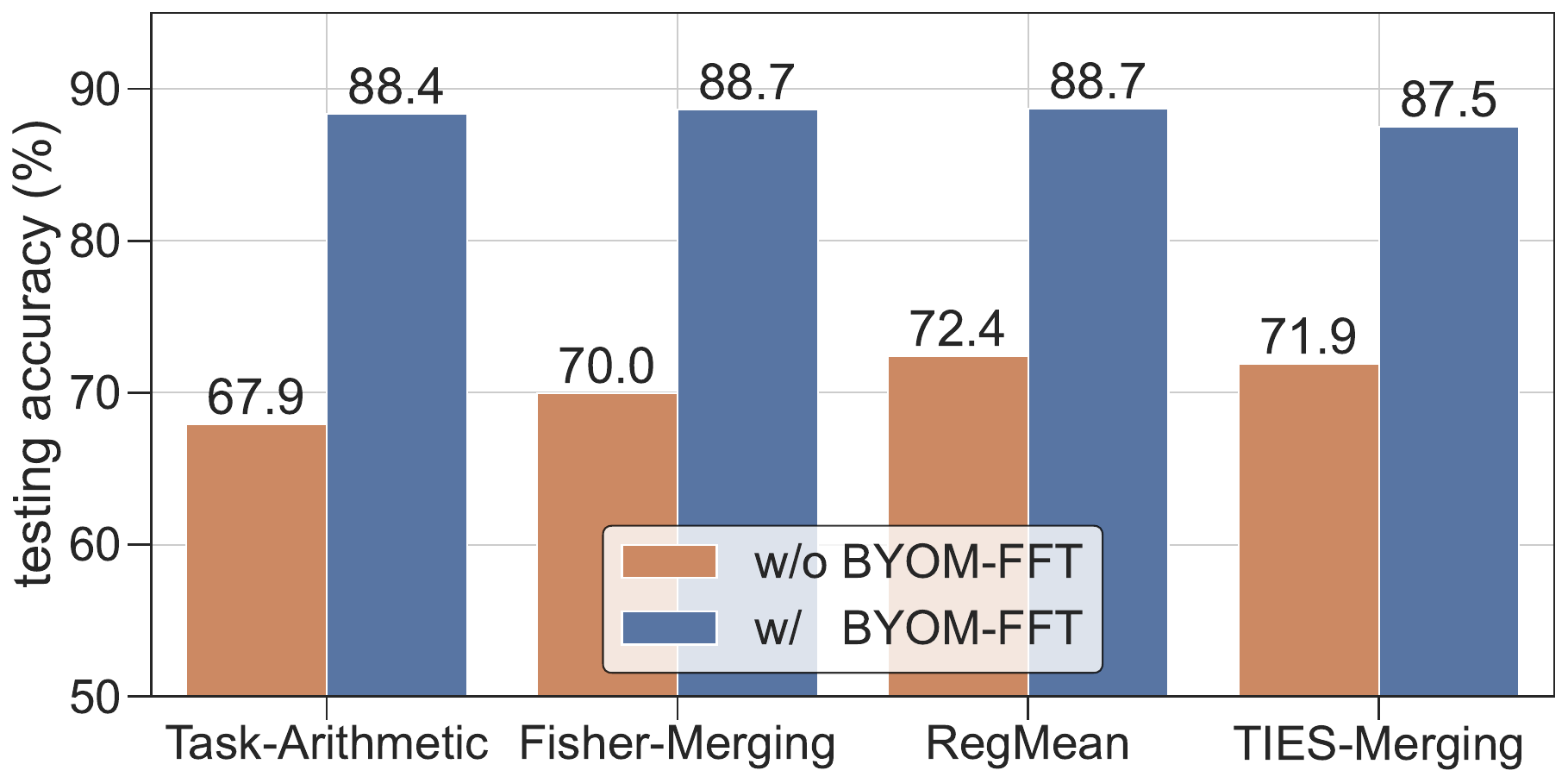}
		\vskip -.2in
		\caption{Effects of integrating BYOM-FFT into existing merging methods.} 
		\label{fig:comb-BYOM-merging}
		\vskip -.2in
	\end{figure}
	
	The proposed BYOM-FFT is general and can be combined with any existing merging method. 
	In Sections \ref{sec:expt-setup} and \ref{sec:exp_nlp}, we use Task-Arithmetic to obtain $\vtheta^\star$.
	In this section, we conduct experiments using the setting with \textit{ViT-B/32} to verify the benefits of integrating BYOM-FFT into other merging methods. 
	Figure \ref{fig:comb-BYOM-merging}
	shows the testing accuracy (detailed results are shown in Table \ref{table:expt-vairant-of-BYOMfft} of Appendix \ref{apd-sec:comb-BYOM}). 
	As can be seen, BYOM-FFT largely boosts the performance of existing methods (i.e., Task-Arithmetic, Fisher-Merging, RegMean, TIES-Merging).

	\begin{figure*}[!h]
		\centering
		% \vskip -.1in
		%		\!\!
		\subfigure[\label{fig:vitb32_sparse_all}\textit{ViT-B/32}.]{\includegraphics[width=0.3\textwidth]{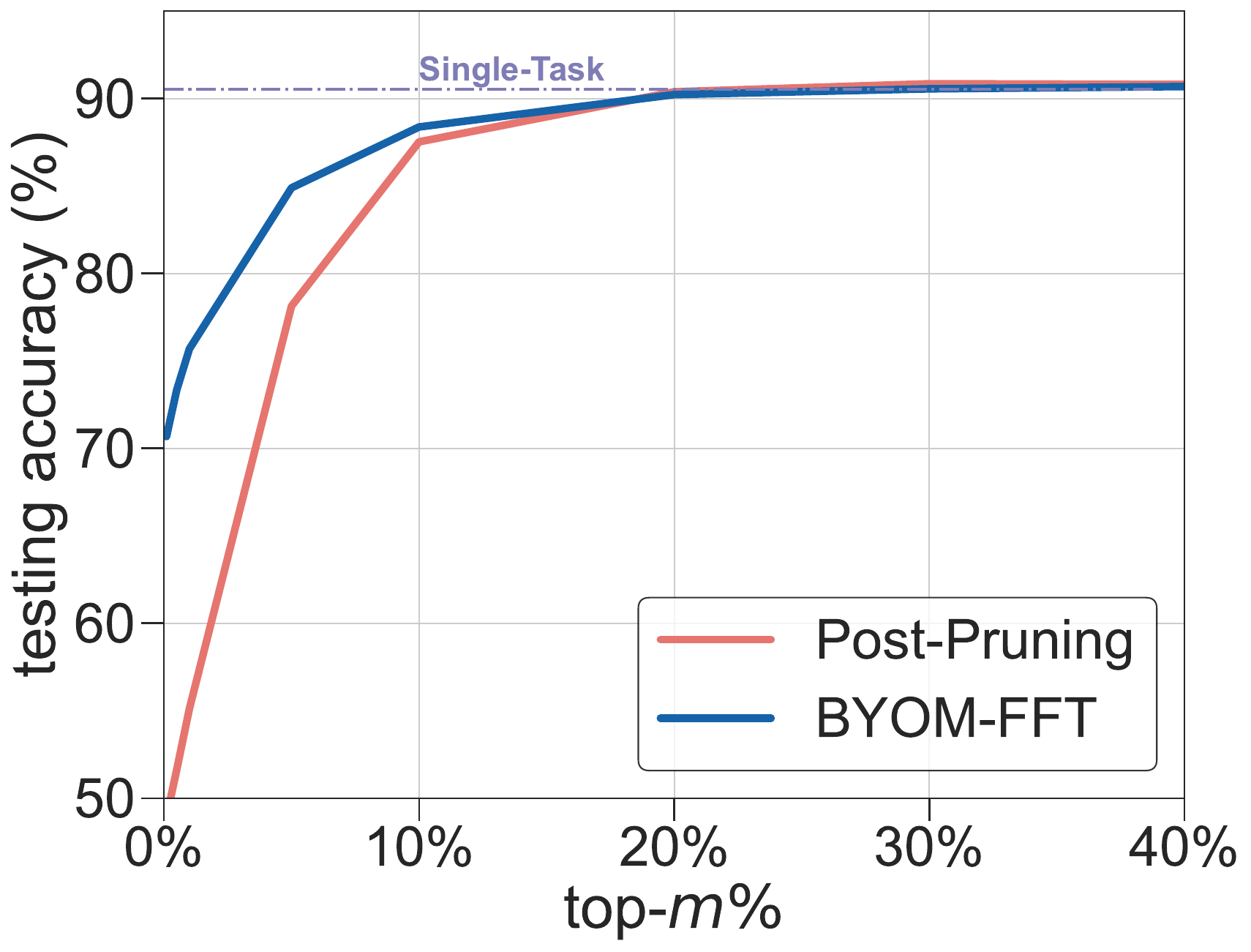}} \; 
		\subfigure[\label{fig:vitb16_sparse_all}\textit{ViT-B/16}.]{\includegraphics[width=0.3\textwidth]{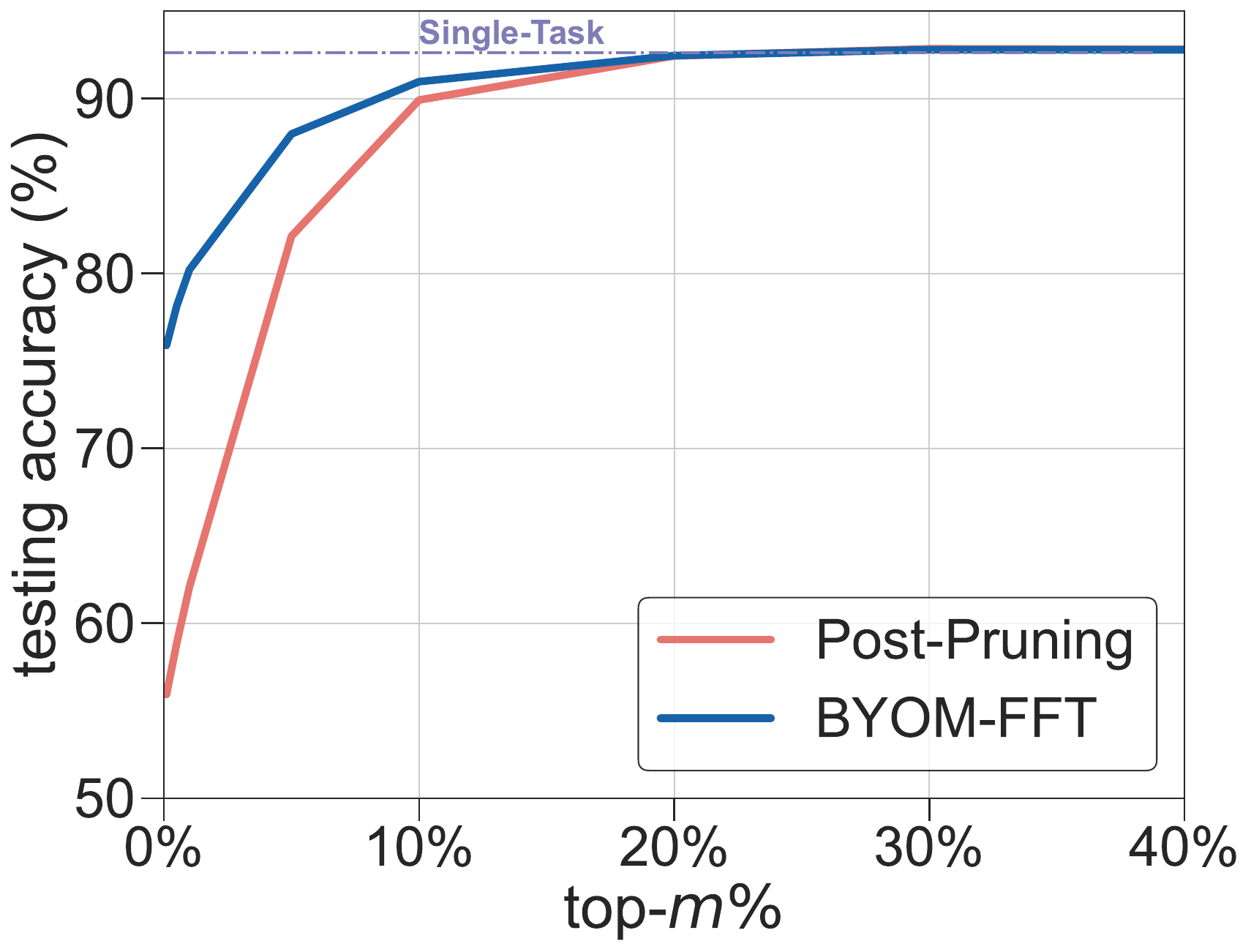}} \; 
		\subfigure[\label{fig:vitL14_sparse_all}\textit{ViT-L/14}.]{\includegraphics[width=0.3\textwidth]{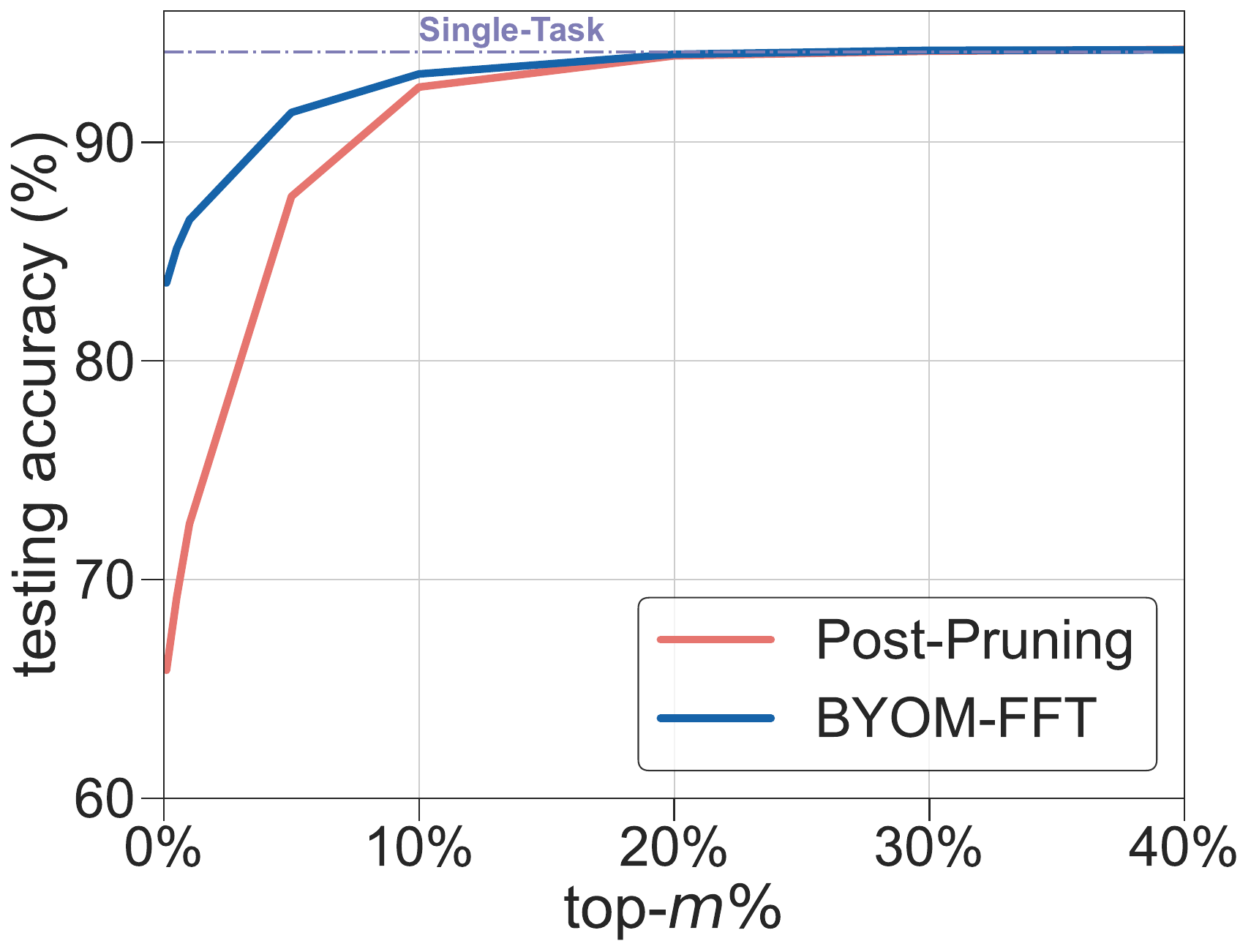}} \; 
		\vskip -.2in
		\caption{Curves of accuracy (averaged over eight tasks) w.r.t. top-$m\%$ values of task vectors.}
		\label{fig:acc-all-sparse}
		\vskip -.1in
	\end{figure*}
	
	\begin{figure*}[!h]
		\centering
		\begin{minipage}{0.3\linewidth}
			\centering
			\includegraphics[width=0.99\textwidth]{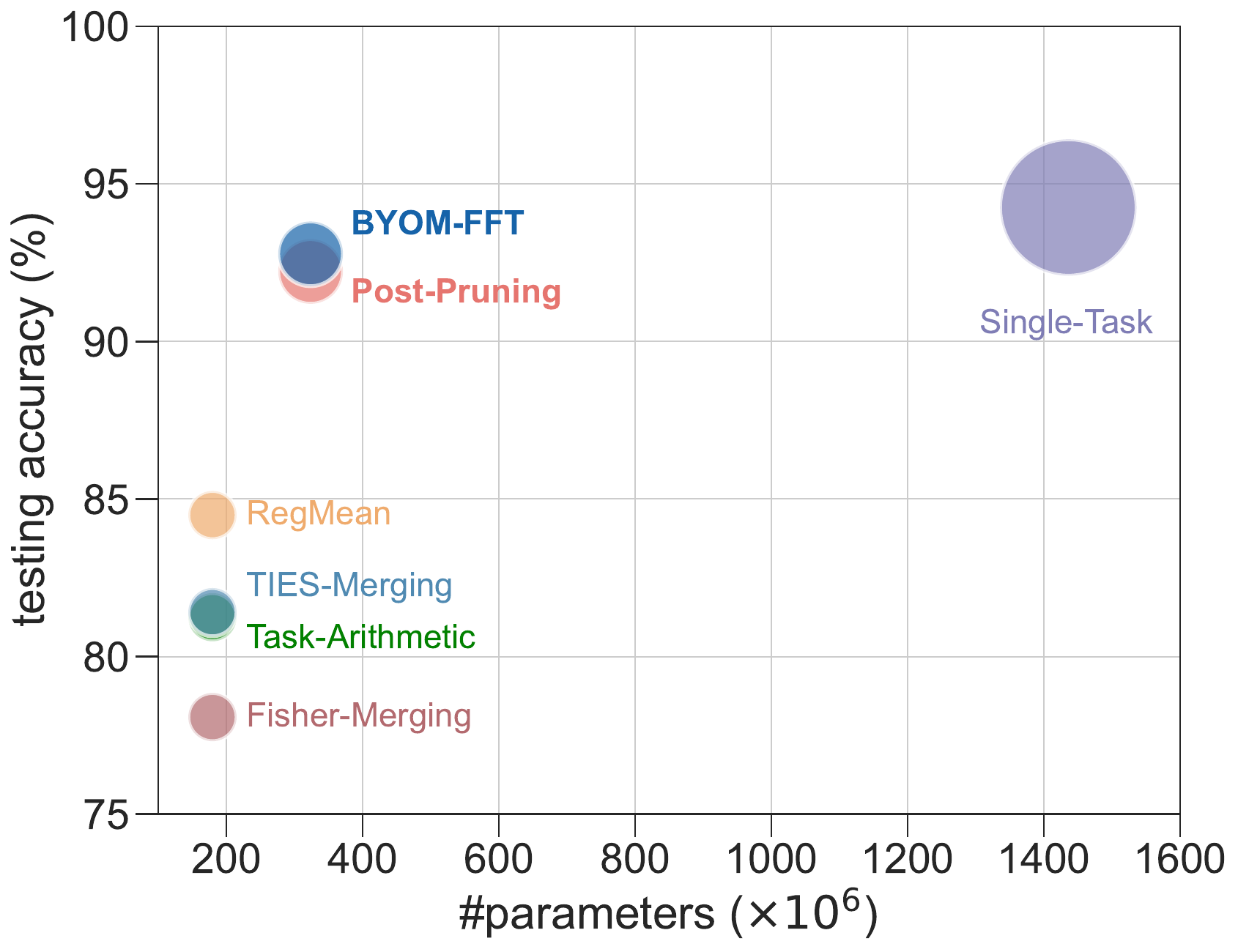}
			\vskip -.2in
			\caption{Testing accuracy (averaged over eight tasks) for merging methods with the \textit{ConvNeXt-Base} network backbone.} 
			\label{fig:fft-convx-bubble}
			\vspace{-1.5em}
		\end{minipage}  \;\;\;
		\begin{minipage}{0.3\linewidth}
			\centering
			\includegraphics[width=0.95\textwidth]{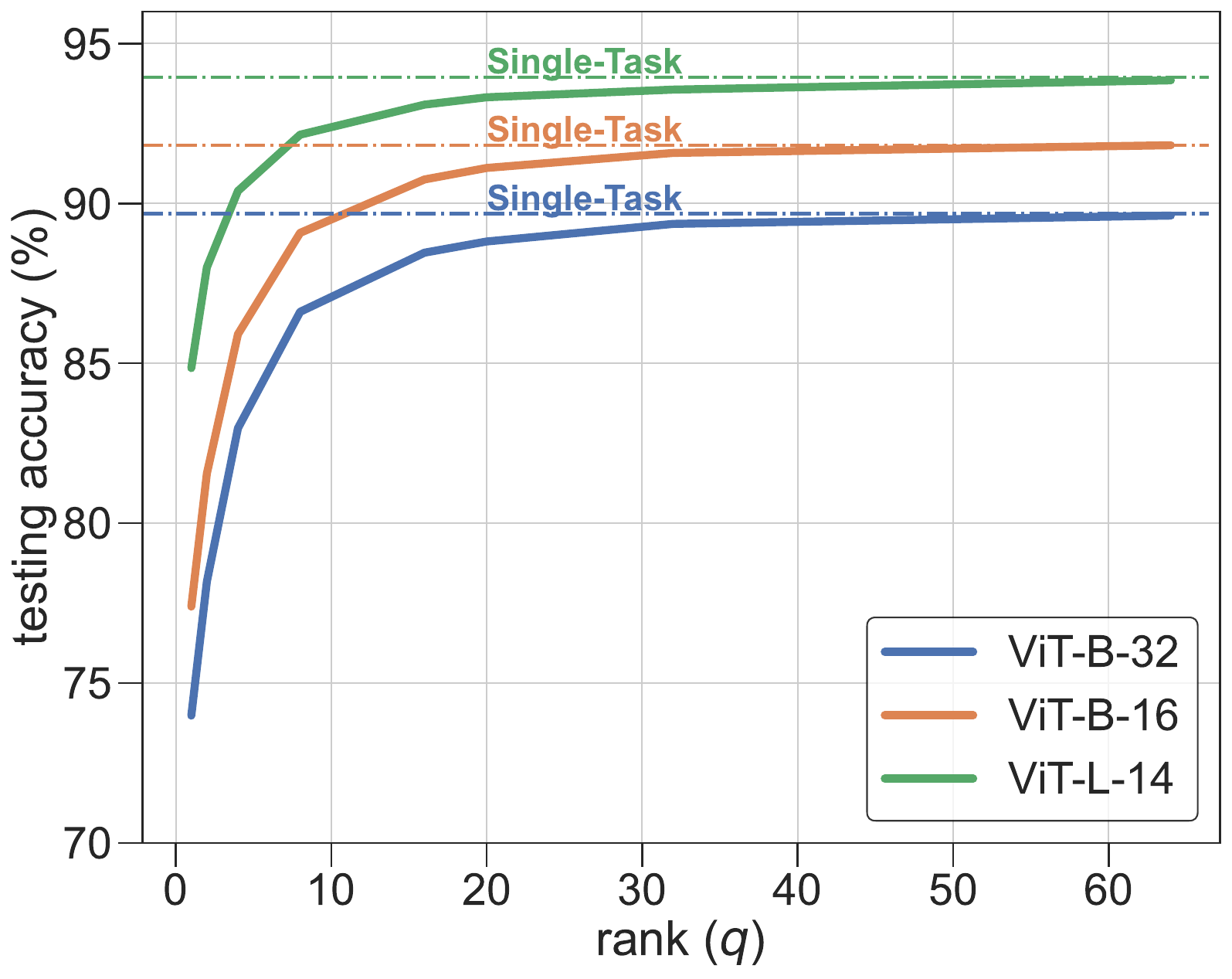}
			\vskip -.2in
			\caption{Curves of average accuracy w.r.t. rank $q$ in BYOM-LoRA.} 
			\label{fig:acc-all-rank}
			\vspace{-1.5em}
		\end{minipage} \;\;\;
		\begin{minipage}{0.3\linewidth}
			\centering
			% \vskip .15in
			\includegraphics[width=0.99\textwidth]{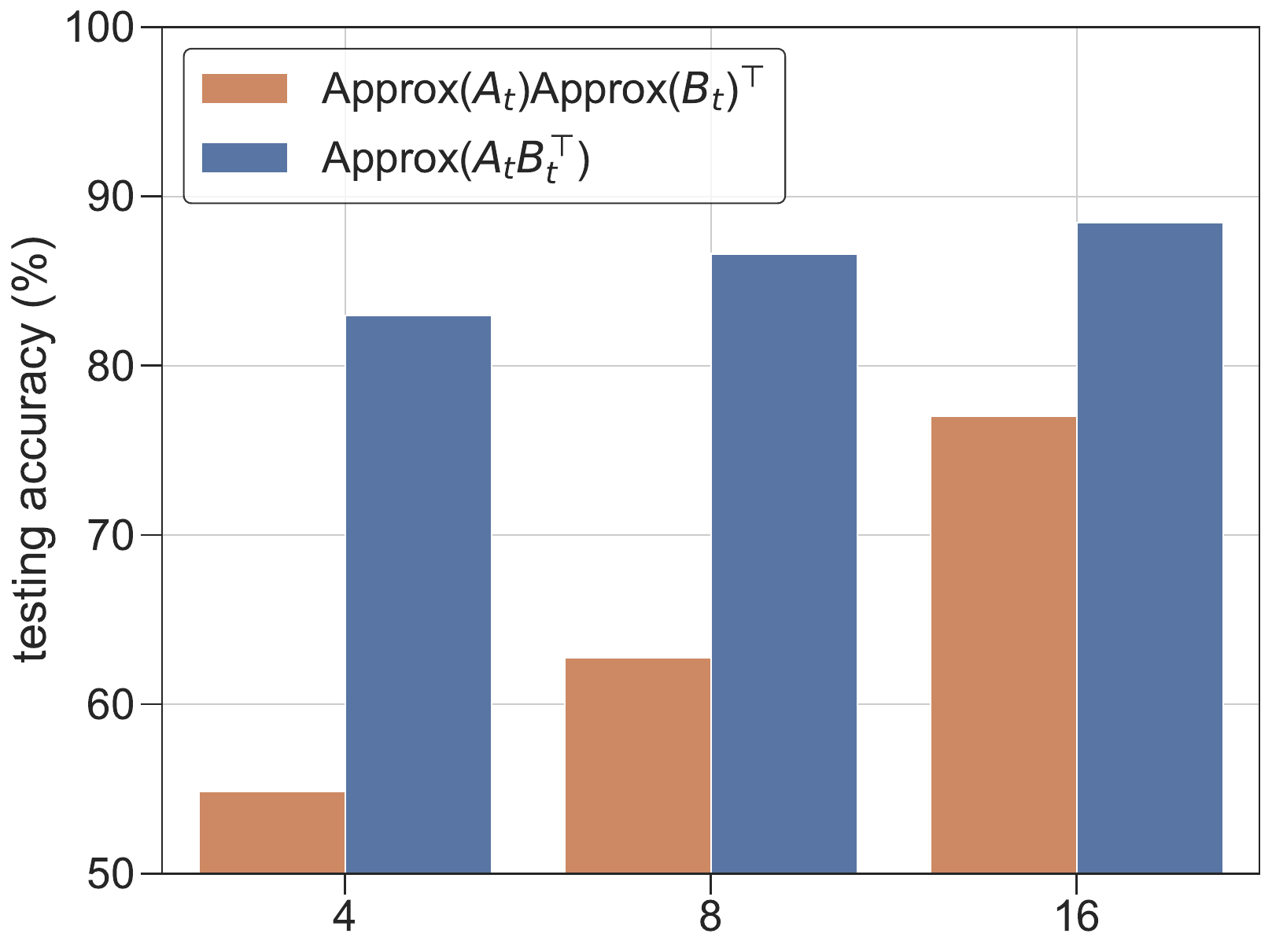}
			\vskip -.2in
			\caption{Ablation study on approximating the LoRA matrices.} 
			\label{fig:abl-lora-svd-ab}
			\vspace{-1.5em}
		\end{minipage} 
	\end{figure*}

	\subsection{Effects of $m\%$ on Post-Pruning and  BYOM-FFT}
	
	In this section, we conduct experiments to study the effects of $m\%$ on the performance of Post-Pruning and BYOM-FFT using the settings in Section \ref{sec:expt-setup}. Figure \ref{fig:acc-all-sparse} shows the testing accuracy (averaged over eight tasks) w.r.t. $m\% \in [0\%, 40\%]$ using \textit{ViT-B/32}, \textit{ViT-B/16}, and \textit{ViT-L/14}. 
	As can be seen, the accuracies of Post-Pruning and BYOM-FFT increase when $m\%$ increases. 
	When $m\%$ is larger than $20\%$, their accuracies reach the Single-Task performance and saturates. 
	As for $m\%\leq 10\%$, BYOM-FFT always performs better than Post-Pruning, suggesting that shared knowledge is useful when pruning most parameters.

	\subsection{Experiments with a CNN-based Model}
	
	As previous experiments are based on Transformers, here we conduct an experiment using a CNN-based model \textit{ConvNeXt-Base}~\citep{liu2022convnet} on eight CV tasks used in Section~\ref{sec:expt-setup}. 
	Figure~\ref{fig:fft-convx-bubble} shows the testing accuracy (detailed results are in Table \ref{table:expt-cnn} of Appendix~\ref{sec:expt-cnn}). 
	As we can see, 
	BYOM-FFT (10\%) performs better than existing merging methods by a large margin,
	showing that task-specific knowledge is beneficial to boosting performance.
	Moreover,
	compared with Single-Task method, BYOM-FFT($10\%$) achieves comparable performance but have $4.5\times$ fewer parameters. 
	Besides,
	BYOM-FFT outperforms Post-Pruning, showing that the shared knowledge is useful.	
	
	\subsection{Effects of $q$ on BYOM-LoRA}
	We conduct experiments to study the effects of rank $q$ on BYOM-LoRA using the settings in Section~\ref{sec:expt-setup}. 
	Figure \ref{fig:acc-all-rank} shows the testing accuracy (averaged over eight tasks) w.r.t. $q$. 
	As can be seen, increasing $q$ leads to better performance. 
	Furthermore, BYOM-LoRA with rank $40$ performs almost the same as Single-Task (LoRA finetuned). 
	Hence, using a lower-rank matrix (e.g., $16$) to approximate the LoRA matrix is parameter-efficient and does not cause significant deterioration in performance.
	
	\subsection{Ablation Study on Approximating LoRA Matrices}
	
	In Section~\ref{sec:lora},
	we perform singular value decomposition to ${\bf A}_t{\bf B}_t^\top$ as it can obtain the best rank-$q$ approximation of ${\bf A}_t{\bf B}_t^\top$, i.e., $\arg\min_{\text{rank}({\bf C})\leq q}\| {\bf C} - {\bf A}_t{\bf B}_t^\top\|_{\text{F}}$. 
	Note that ${\bf A}_t{\bf B}_t^\top$ is applied to the pre-trained weight $\theta_0$ (i.e., $\vtheta_t=\vtheta_0+{\bf A}_t{\bf B}_t^\top$), thus, approximating ${\bf A}_t{\bf B}_t^\top$ might be more effective than approximating ${\bf A}_t$ and ${\bf B}_t$ separately. 
	We conduct an ablation experiment with \textit{ViT-B/32} using the setting in Section~\ref{sec:expt-setup}. 
	Figure~\ref{fig:abl-lora-svd-ab} compares the testing accuracy of $\vtheta_0+\text{Approx}({\bf A}_t{\bf B}_t^\top)$ with $\vtheta_0 + \text{Approx}({\bf A}_t)\text{Approx}({\bf B}_t)^\top$ (detailed results are in Table~\ref{table:expt-lora-ab} of Appendix~\ref{sec:expt-abl-lora}). 
	As shown, Approx(${\bf A}_t{\bf B}_t^\top$) performs better.
	
	%%%%%%%%%%%%%%%%%%%%%%%%%%%%%%%%%
	%% Section: Conclusion
	%%%%%%%%%%%%%%%%%%%%%%%%%%%%%%%%%
	
	\section{Conclusion}
	
	In this paper, we studied the problem of building a multi-task model from task-specific finetuned models.
	We proposed two parameter-efficient methods:
	(i)
	BYOM-FFT for fully finetuned models
	by compressing task-specific knowledge into a sparse vector;
	and  (ii)
	BYOM-LoRA for LoRA finetuned models
	by compressing the LoRA matrices.
	Both methods are data-free and computation-efficient in merging.
	Extensive experiments on CV and NLP tasks demonstrate that BYOM-FFT and BYOM-LoRA significantly outperform existing merging methods. 
	Additionally, the proposed methods achieve comparable performance to the Single-Task method but are much more
	parameter-efficient.
	Moreover, BYOM-FFT is general and can be combined with any existing merging algorithms to boost performance.
	
	%\newpage
	
	\section*{Broader Impact and Ethics Statements}
	This paper presents a novel method for building multi-task models from task-specific finetuned models. 
	Our goal is to advance the field of machine learning. 
	Our work improves the efficiency and accuracy of multi-task models, which can have practical applications in various fields.
	As merging task-specific models does not require user data, 
	there is no concern about ethical considerations and data privacy.

	\bibliography{paper_arxiv}
	\bibliographystyle{icml2024}
	
	%%%%%%%%%%%%%%%%%%%%%%%%%%%%%%%%%
	%% Section: Appendix
	%%%%%%%%%%%%%%%%%%%%%%%%%%%%%%%%%
	\appendix
	\onecolumn
	
	\section*{Limitations and Future Works}
	
	Like existing merging methods (Task-Arithmetic, Fisher-Merging, RegMean, TIES-Merging), the proposed BYOM is only suitable for task-specific models with the same architecture. Another limitation is that BYOM requires the task-specific models to be finetuned from the same pre-trained model, which is also necessary for existing merging methods. 
	Existing merging methods and BYOM need to know the specification of dataset sources when addressing different classification tasks.
	Weakening these assumptions is an important research direction in merging models.

	\section{Additional Experiments}
	
	\subsection{Experimental Results using \textit{ViT-B/16}}
	\label{sec:expt-vit-b16}
	
	Table~\ref{table:vitb16} shows the testing accuracy on eight image classification tasks using \textit{ViT-B/16}.
	As can be seen,
	for merging fully finetuned models, 
	BYOM-FFT performs much better than existing merging methods.
	Furthermore, 
	BYOM-FFT achieves comparable performance to Single-Task but is more parameter-efficient.
	For reusing LoRA finetuned models,
	BYOM-LoRA outperforms existing merging methods by a large margin.

	\begin{table*}[!h]
		% \vskip -.1in
		\centering
		\caption{Testing accuracy on eight CV task  using \textit{ViT-B/16}.
		} 
		\label{table:vitb16}
		\resizebox{.8\textwidth}{!}{
			\renewcommand{\arraystretch}{1.15}
			\begin{NiceTabular}{c@{\hspace{.05in}}l@{\hspace{-.15in}}c@{\hspace{.08in}}c@{\hspace{.08in}}c@{\hspace{.08in}}c@{\hspace{.08in}}c@{\hspace{.08in}}c@{\hspace{.08in}}c@{\hspace{.08in}}c@{\hspace{.1in}}c@{\hspace{.1in}}c}
				\CodeBefore  
				\rectanglecolor{blue3!5}{3-1}{13-1}
				\rectanglecolor{blue3!10}{14-1}{21-1}
				\rectanglecolor{Gray}{11-2}{13-12}
				\rectanglecolor{Gray}{19-2}{21-12}
				\Body
				\arrayrulecolor{black}\specialrule{1.3pt}{.3\jot}{0.3pc}
				& & \#params (M)
				& \textit{MNI} & \textit{GTS} & \textit{SVH} & \textit{RES} & \textit{SUN} & \textit{EUR} & \textit{DTD} & \textit{CAR} & Avg \\
				\arrayrulecolor{black}\specialrule{1.3pt}{.3\jot}{0.3pc}
				%%%%
				%%%% fll ft
				%%%%
				& Pre-Trained & 112 & 51.79 & 43.34 & 51.98 & 65.76 & 65.50 & 55.22 & 45.11 & 64.57 & 55.41 \\
				\arrayrulecolor{black}\specialrule{1.3pt}{.3\jot}{0.3pc}
				\multirow{12}{*}{\STAB{\rotatebox[origin=c]{90}{Fully FT}}}
				&Single-Task & 894 & 99.72 & 99.15 & 97.86 & 96.57 & 78.71 & 99.33 & 82.29 & 87.20 & 92.60 \\
				\cmidrule{2-12}
				&Task-Arithmetic & 112 & 97.35 & 71.39 & 80.50 & 75.71 & 67.88 & 82.63 & 52.34 & 70.74 & 74.82 \\
				&Fisher-Merging & 112 & 94.52 & 61.21 & 73.24 & 75.25 & 68.54 & 80.41 & 50.74 & 69.94 & 71.73 \\
				&RegMean & 112 & 96.93 & 70.26 & 83.79 & 77.60 & 69.10 & 88.85 & 54.63 & 71.67 & 76.60 \\
				&TIES-Merging & 112 & 98.75 & 74.43 & 88.84 & 78.48 & 66.21 & 85.93 & 57.13 & 73.15 & 77.86 \\
				\cmidrule{2-12}
				&Post-Pruning ($1\%$) & 121 & 60.94 & 47.66 & 60.54 & 73.97 & 68.52 & 66.15 & 49.63 & 69.29 & 62.09 \\
				&Post-Pruning ($5\%$) & 157 & 96.06 & 77.36 & 82.08 & 88.70 & 74.42 & 94.22 & 64.89 & 79.28 & 82.13 \\
				&Post-Pruning ($10\%$) & 201 & 99.32 & 94.83 & 94.43 & 94.62 & 77.00 & 98.44 & 76.01 & 84.62 & 89.91 \\
				\cmidrule{2-12}
				&BYOM-FFT ($1\%$) & 121 & 98.32 & 79.85 & 85.12 & 82.89 & 71.22 & 89.30 & 59.79 & 75.33 & 80.23 \\
				&BYOM-FFT ($5\%$) & 157 & 99.38 & 92.91 & 93.90 & 92.60 & 74.99 & 97.11 & 71.12 & 81.72 & 87.97 \\
				&BYOM-FFT ($10\%$) & 201 & \textbf{99.56} & \textbf{97.34} & \textbf{96.91} & \textbf{95.30} & \textbf{77.11} & \textbf{98.67} & \textbf{77.77} & \textbf{85.04} & \textbf{90.96} \\
				\arrayrulecolor{black}\specialrule{1.3pt}{.3\jot}{0.3pc}
				%%%%
				%%%% lora ft
				%%%%
				\multirow{8}{*}{\STAB{\rotatebox[origin=c]{90}{LoRA FT}}}
				&Single-Task & 192 & 99.77 & 99.11 & 97.72 & 96.21 & 76.63 & 98.89 & 79.95 & 86.27 & 91.82 \\
				\cmidrule{2-12}
				&Task-Arithmetic & 112 & 95.59 & 63.06 & 77.30 & 72.92 & 66.05 & 82.67 & 49.04 & 64.46 & 71.38 \\
				& Fisher-Merging & 112 & 94.51 & 61.19 & 73.22 & 75.24 & 68.57 & 80.41 & 50.74 & 69.93 & 71.73 \\
				&RegMean & 112 & 97.89 & 68.73 & 85.26 & 76.30 & 68.17 & 91.96 & 52.66 & 70.54 & 76.44 \\
				&TIES-Merging & 112 & 90.69 & 54.52 & 71.18 & 74.41 & 68.02 & 77.59 & 48.56 & 67.98 & 69.12 \\
				\cmidrule{2-12}
				&BYOM-LoRA (4) &  114 & 99.35 & 93.96 & 95.52 & 88.65 & 72.21 & 96.81 & 69.73 & 71.05 & 85.91 \\
				&BYOM-LoRA (8) & 117 & 99.64 & 97.51 & 97.16 & 93.40 & 73.55 & 98.52 & 76.12 & 76.72 & 89.08 \\
				&BYOM-LoRA (16) & 122 & \textbf{99.66} & \textbf{98.54 }& \textbf{97.61} & \textbf{95.25} & \textbf{75.54} & \textbf{98.78} & \textbf{78.72} & \textbf{81.88} & \textbf{90.75} \\
				\arrayrulecolor{black}\specialrule{1.3pt}{.3\jot}{0.3pc}
		\end{NiceTabular}}
		\vskip -.2in
	\end{table*}
	
	\subsection{Combing BYOM-FFT with Existing Merging Models Methods}
	\label{apd-sec:comb-BYOM}
	
	The proposed BYOM-FFT is general and can be combined with any existing merging methods. 
	We conduct experiments using the setting with \textit{ViT-B/32}
	to verify the compatibility of BYOM-FFT with existing merging methods. 
	Table \ref{table:expt-vairant-of-BYOMfft} shows the detailed testing accuracy. 
	As we can see, BYOM-FFT consistently benefits existing methods (Task-Arithmetic, Fisher-Merging, RegMean, TIES-Merging).
	
	\begin{table}[t]
		% \vskip -.2in
		\caption{Accuracy on eight tasks with \textit{ViT-B/32} when combing the proposed BYOM-FFT with existing merging methods.}
		\label{table:expt-vairant-of-BYOMfft}
		\renewcommand{\arraystretch}{1.15}
		\resizebox{.98\textwidth}{!}{
			\begin{NiceTabular}{lcccccccccc}
				\toprule
				\text{Methods} & \text{\#params (M)} & \textit{MNI} & \textit{GTS} & \textit{SVH} & \textit{RES} & \textit{SUN} & \textit{EUR} & \textit{DTD} & \textit{CAR} & \text{Avg} \\
				\midrule
				{Single-Task}  & {908} &{ 99.72} & {99.23} & { 97.42} & {95.56} & {75.03} & {99.00} & {79.47} & {78.73} & {90.52} \\
				\midrule
				\text{Task-Arithmetic} & 113 & 93.27 & 65.99 & 71.62 & 71.57 & 63.63 & 78.41 & 51.76 & 61.50 & 69.72 \\ \rowcolor{Gray}
				\text{Task-Arithmetic + BYOM-FFT } (1\%) & 123 & 96.17 & 76.33 & 79.27 & 78.03 & 66.88 & 84.89 & 58.03 & 65.99 & 75.70 \\\rowcolor{Gray}
				\text{Task-Arithmetic + BYOM-FFT } (5\%) & 159 & 99.12 & 92.66 & 91.86 & 88.48 & 71.35 & 94.85 & 67.77 & 73.08 & 84.90 \\\rowcolor{Gray}
				\text{Task-Arithmetic + BYOM-FFT } (10\%) & 204 &  {\bf 99.49} &  {\bf 97.57} & {\bf 95.92} & {\bf 93.00} & {\bf 73.52} & {\bf 97.63} & {\bf 72.98} & {\bf 76.92} & {\bf 88.38} \\
				\midrule
				\text{Fisher-Merging} & 113 & 80.71 & 75.15 & 74.08 & 70.24 & 65.25 & 81.48 & 49.84 & 62.90 & 69.96 \\\rowcolor{Gray}
				\text{Fisher-Merging + BYOM-FFT } (1\%) & 123 & 92.29 & 69.23 & 71.50 & 77.54 & 68.27 & 79.59 & 56.28 & 67.40 & 72.76\\\rowcolor{Gray}
				\text{Fisher-Merging + BYOM-FFT } (5\%) & 159 & 98.81 & 91.39 & 90.31 & 88.73 & 72.38 & 94.33 & 67.18 & 74.26 & 84.67 \\\rowcolor{Gray}
				\text{Fisher-Merging + BYOM-FFT } (10\%) & 204 & {\bf 99.46} & {\bf 97.26} & {\bf 95.73} & {\bf 93.46} & {\bf 74.43} & {\bf 97.37} & {\bf 73.72} & {\bf 77.81} & {\bf 88.66} \\
				\midrule
				\text{RegMean} & 113 & 92.55 & 65.12 & 75.48 & 75.56 & 65.72 & 84.33 & 56.01 & 64.54 & 72.41 \\\rowcolor{Gray}
				\text{RegMean + BYOM-FFT } (1\%) & 123 & 93.79 & 71.46 & 78.77 & 77.95 & 67.47 & 87.15 & 58.14 & 66.47 & 75.15 \\\rowcolor{Gray}
				\text{RegMean + BYOM-FFT } (5\%) & 159 & 98.59 & 91.83 & 91.74 & 88.65 & 72.01 & 96.19 & 67.98 & 73.77 & 85.10\\\rowcolor{Gray}
				\text{RegMean + BYOM-FFT } (10\%) & 204 & {\bf 99.37} & {\bf 97.39} & {\bf 95.76} & {\bf 93.56} & {\bf 74.37} & {\bf 97.89} & {\bf 74.04} & {\bf 77.09} & {\bf 88.68}\\
				\midrule
				\text{TIES-Merging} & 113 & 97.79 & 75.30 & 84.10 & 70.71 & 59.24 & 75.89 & 53.51 & 58.72 & 71.91 \\\rowcolor{Gray}
				\text{TIES-Merging + BYOM-FFT} (1\%) & 123 &  98.82 & 86.25 & 87.27 & 75.79 & 61.29 & 87.15 & 58.78 & 62.22 & 77.20  \\\rowcolor{Gray}
				\text{TIES-Merging + BYOM-FFT} (5\%) & 159 & 99.44 & 96.42 & 94.52 & 86.86 & 67.01 & 95.22 & 67.45 & 70.41 & 84.67\\\rowcolor{Gray}
				\text{TIES-Merging + BYOM-FFT} (10\%) & 204 & {\bf 99.65} & {\bf 98.22} & {\bf 96.40} & {\bf 91.86} & {\bf 70.32} & {\bf 97.22} & {\bf 72.45} & {\bf 74.12} & {\bf 87.53}\\
				\bottomrule
			\end{NiceTabular}
		} 
		% \vskip .1in
	\end{table}

	\subsection{Ablation Study on Approximating LoRA Matrices}
	\label{sec:expt-abl-lora}
	
	We perform singular value decomposition to ${\bf A}_t{\bf B}_t^\top$ as it can obtain the best rank-$k$ approximation of ${\bf A}_t{\bf B}_t^\top$, i.e., $\arg\min_{\text{rank}({\bf C})\leq k}\| {\bf C} - {\bf A}_t{\bf B}_t^\top\|_{\text{F}}$. 
	Note that ${\bf A}_t{\bf B}_t^\top$ is applied to the pre-trained weight $\theta_0$ (i.e., $\vtheta_t=\vtheta_0+{\bf A}_t{\bf B}_t^\top$), thus, approximating ${\bf A}_t{\bf B}_t^\top$ might be more effective than approximating ${\bf A}_t$ and ${\bf B}_t$ separately. 
	We conduct an ablation experiment with \textit{ViT-B/32} using the setting in Section \ref{sec:expt-setup}. 
	Table \ref{table:expt-lora-ab} shows the testing accuracy of $\vtheta_0+\text{Approx}({\bf A}_t{\bf B}_t^\top)$ with $\vtheta_0 + \text{Approx}({\bf A}_t)\text{Approx}({\bf B}_t)^\top$. 
	As can be seen, Approx(${\bf A}_t{\bf B}_t^\top$) consistently performs better than $\text{Approx}({\bf A}_t)\text{Approx}({\bf B}_t)^\top$.
	
	\begin{table}[t]
		% \vskip -.25in
		\centering
		\caption{Accuracy on eight tasks of Approx($\vA_t$){Approx}$(\vB_t)^\top $ and \text{Approx($\vA_t\vB_t^\top$)}  when reusing LoRA finetuned models with \textit{ViT-B/32}.}
		\label{table:expt-lora-ab}
		\renewcommand{\arraystretch}{1.15}
		\resizebox{.85\textwidth}{!}{
			\begin{NiceTabular}{lcccccccccc}
				\toprule
				& \text{rank}  & \textit{MNI} & \textit{GTS} & \textit{SVH} & \textit{RES} & \textit{SUN} & \textit{EUR} & \textit{DTD} & \textit{CAR} & \text{Avg} \\
				\midrule
				Approx($\vA_t$){Approx}$(\vB_t)^\top $ & 4  & 61.42 & 38.29 & 39.17 & 64.41 & 64.35 & 64.74 & 45.85 & 60.63 & 54.86 \\
				\text{Approx($\vA_t\vB_t^\top$)} & 4  & {\bf 99.16} & {\bf 92.04} & {\bf 93.98} & {\bf 86.48} & {\bf 68.61} & {\bf 95.37} & {\bf 65.37} & {\bf 62.74} & {\bf 82.97} \\
				\midrule
				Approx($\vA_t$){Approx}$(\vB_t)^\top $& 8   & 79.39 & 45.81 & 50.33 & 70.73 & 65.50 & 79.67 & 48.35 & 62.09 & 62.73 \\\rowcolor{Gray}
				\text{Approx(${\bf A}_t{\bf B}_t^\top$)} & 8 & {\bf 99.54} & {\bf 96.23} & {\bf 96.45} & {\bf 92.16} & {\bf 70.33} & {\bf 98.26} & {\bf 72.55} & {\bf 67.35} & {\bf 86.61} \\
				\midrule
				Approx($\vA_t$){Approx}$(\vB_t)^\top $& 16   & 96.25 & 72.49 & 81.47 & 81.67 & 67.40 & 93.56 & 57.18 & 66.22 & 77.03\\\rowcolor{Gray}
				\text{Approx(${\bf A}_t{\bf B}_t^\top$)} & 16   & {\bf 99.62} & {\bf 97.99} & {\bf 97.08} & {\bf 94.56} & {\bf 72.29} &  {\bf 98.37} &  {\bf 76.44} & {\bf 71.31} & {\bf 88.46} \\
				\bottomrule
			\end{NiceTabular}
		} 
		% \vskip -.2in
	\end{table}
	
	\subsection{Experiments on the \textit{VTAB} Dataset}
	
	We conduct an experiment on the \textit{Natural} group of \textit{VTAB} \citep{zhai2019large} using \textit{ViT-B/16},
	where tasks are more similar to each other.  Table \ref{table:expt-vtab} shows the testing accuracy. The observations are consistent with the experimental results in Section \ref{sec:expt-setup}. 
	Specifically, as we can see, by keeping top-10\% values, both BYOM-FFT and Post-Pruning achieve comparable performance with Single-Task, but are more parameter-efﬁcient (4.5$\times$ fewer parameters). BYOM-FFT  performs better than the existing merging method by a large margin, showing the effectiveness of introducing sparse task-speciﬁc vectors into the merged model.
	Compared with Post-Pruning, BYOM-FFT achieves higher accuracy (averaged over seven tasks), showing that merging the task-speciﬁc models before pruning the task vectors is more effective.
	
	\begin{table}[!t]
		\centering
		% \vskip -.2in
		\caption{\color{rev} Accuracy on seven tasks from the \textit{Natural} group of \textit{VTAB}.}
		\label{table:expt-vtab}
		\renewcommand{\arraystretch}{1.15}
		\resizebox{.85\textwidth}{!}{
			\begin{NiceTabular}{lccccccccc}
				\toprule
				& \text{\#params (M)}
				& \textit{CIF} & \textit{CAL} & \textit{DTD} & \textit{FLO} & \textit{PET} & \textit{SVH} & \textit{SUN} & \text{Avg} \\
				\midrule
				\text{Pre-Trained} & 112 & 66.91 & 82.76 & 45.11 & 71.33 & 87.19 & 51.98 & 65.50 & 67.25 \\
				\midrule
				\text{Single-Task} & 894 & 90.23 & 97.20 & 82.29 & 94.88 & 94.55 & 97.86 & 78.71 & 90.82  \\
				\midrule
				\text{Task-Arithmetic} & 112 & 83.33 & 87.07 & 52.87 & 66.87 & 89.10 & 83.52 & 67.01 & 75.68  \\
				\text{Fisher-Merging} &112 & 80.81 & 86.96 & 51.28 & 73.72 & 89.51 & 69.92 & 69.32 & 74.50 \\
				\text{RegMean} & 112 & 81.32 & 87.07 & 55.53 & 75.41 & 90.38 & 84.94 & 69.88 & 77.79 \\
				\text{TIES-Merging} & 112 & 82.77 & 87.69 & 57.39 & 70.39 & 89.59 & 88.28 & 67.42 & 77.65 \\
				\midrule
				\text{Post-Pruning ($1\%$)} & 121 & 74.37 & 85.06 & 49.63 & 73.59 & 88.23 & 60.53 & 68.51 & 71.42 \\
				\text{Post-Pruning ($5\%$)} & 157 & 86.55 & 90.60 & 64.89 & 78.45 & 91.58 & 82.06 & 74.41 & 81.22 \\
				\text{Post-Pruning ($10\%$)} & 201& 89.61 & 93.68 & 76.01 & {\bf 84.34} & {\bf 94.41} & 94.42 & {\bf 77.00} & 87.07 \\
				\midrule \rowcolor{Gray}
				\text{BYOM-FFT ($1\%$)} & 121& 85.39 & 89.93 & 58.30 & 70.89 & 91.61 & 87.78 & 69.76 & 79.09 \\\rowcolor{Gray}
				\text{BYOM-FFT ($5\%$)} & 157& 88.50 & 92.56 & 69.73 & 77.62 & 93.59 & 94.71 & 74.18 & 84.41 \\\rowcolor{Gray}
				\text{BYOM-FFT ($10\%$)} &201 & {\bf 89.75} & {\bf 93.90} & {\bf 76.22} & 83.98 & 94.06 & {\bf 97.01} & 76.46 & {\bf 87.34} \\
				\bottomrule
			\end{NiceTabular}
		}  
	\end{table}
	
	\newpage
	\subsection{Experiments using CNN-based Models}
	\label{sec:expt-cnn}
	
	Table \ref{table:expt-cnn}
	shows the testing accuracy on eight image classification tasks using the \textit{ConvNeXt-Base} backbone.
	As can be seen,
	BYOM-FFT achieves comparable performance to Single-Task method but is much more parameter-efficient.
	Moreover, BYOM-FFT significantly outperforms existing 
	merging methods.
	
	\begin{table}[!t]
		% \vskip -.2in
		\centering
		\caption{Accuracy on eight tasks reusing fully finetuned models using \textit{ConvNeXt-Base}.}
		\label{table:expt-cnn}
		\renewcommand{\arraystretch}{1.15}
		\resizebox{.85\textwidth}{!}{
			\begin{NiceTabular}{lcccccccccc}
				\toprule
				& \text{\#params (M)}
				& \textit{MNI} & \textit{GTS} & \textit{SVH} & \textit{RES} & \textit{SUN} & \textit{EUR} & \textit{DTD} & \textit{CAR} & {Avg} \\
				\midrule
				\text{Pre-Trained} & 179 & 64.39 & 46.56 & 53.73 & 65.94 & 71.61 & 52.37 & 61.54 & 91.24 & 63.42\\
				\midrule
				\text{Single-Task}  & 1,435 & 99.78 & 99.22 & 98.01 & 96.67 & 80.49 & 99.19 & 85.74 & 94.91 & 94.25\\
				\midrule
				\text{Task-Arithmetic} & 179 & 97.73 & 81.31 & 82.96 & 76.56 & 72.12 & 78.07 & 68.40 & 92.87 & 81.25\\
				\text{Fisher-Merging} & 179 & 95.47 & 67.67 & 77.93 & 76.21 & 72.80 & 74.22 & 67.82 & 92.53 & 78.08\\
				\text{RegMean} & 179 & 97.92 & 81.25 & 86.47 & 80.65 & 74.00 & 89.26 & 72.55 & 93.82 & 84.49\\
				\text{TIES-Merging} & 179 & 99.16 & 85.32 & 88.83 & 72.97 & 69.44 & 78.37 & 65.27 & 91.93 & 81.41\\
				\midrule
				\text{Post-Pruning ($1\%$)} & 194 & 85.64 & 54.60 & 65.75 & 73.29 & 73.68 & 65.30 & 67.13 & 92.54 & 72.24\\
				\text{Post-Pruning ($5\%$)} & 251 & 99.04 & 83.80 & 89.97 & 87.43 & 77.26 & 91.11 & 76.33 & 94.28 & 87.40\\
				\text{Post-Pruning ($10\%$)} & 323 & 99.65 & 95.65 & 96.47 & 93.21 & {\bf 79.29} & {\bf 97.89} & 80.74 & 94.86 & 92.22\\
				\midrule\rowcolor{Gray}
				\text{BYOM-FFT ($1\%$)} & 194 & 99.02 & 88.49 & 87.86 & 81.89 & 73.79 & 86.26 & 72.07 & 93.56 & 85.37\\\rowcolor{Gray}
				\text{BYOM-FFT ($5\%$)} & 251 & 99.60 & 95.90 & 95.01 & 90.08 & 76.57 & 95.00 & 78.09 & 94.68 & 90.62\\\rowcolor{Gray}
				\text{BYOM-FFT ($10\%$)} & 323 & {\bf 99.72} & {\bf 98.02} & {\bf 97.33} & {\bf 93.65} & 78.66 & 97.41 & {\bf 82.39} & {\bf 95.04} & {\bf 92.78} \\
				\bottomrule
			\end{NiceTabular}
		}  
	\end{table}

	\subsection{Ablation Study on Pruning Methods}
	
	We conduct additional experiments with \textit{ViT-B/32} to compare the performance of pruning $\vtheta_t$ and pruning ${\bf v}_t$ or ${\bf u}_t$. 
	Table \ref{table:ablation-pruning} shows the testing accuracy, where Pruning $\vtheta_t$ (m\%) denotes keeping the top-m\% parameters in $\vtheta_t$. 
	As can be seen, pruning $\vtheta_t$ is not effective. For example, Pruning $\vtheta_t$ (50\%) has very low accuracy. 
	In contrast, keeping top-10\% of ${\bf v}_t$ or ${\bf u}_t$ perform much better (+80\%). 
	Compared with Pruning $\vtheta_t$ (90\%), BYOM-FFT (10\%) achieves comparable performance but has $4\times$ fewer parameters. 
	Hence, pruning ${\bf u}_t$ is more effective and promising than pruning $\vtheta_t$.
	
	\begin{table}[!h]
		% \vskip -.2in
		\centering
		\caption{\color{rev}Comparison between pruning $\vtheta_t$ and pruning ${\bf v}_t$ or $\vu_t$.}
		\label{table:ablation-pruning}
		\renewcommand{\arraystretch}{1.15}
		\resizebox{.85\textwidth}{!}{
			\begin{NiceTabular}{lcccccccccc}
				\toprule
				& \text{\#params (M)} & \textit{MNI} & \textit{GTS} & \textit{SVH} & \textit{RES} & \textit{SUN} & \textit{EUR} & \textit{DTD} & \textit{CAR} & \text{Avg} \\
				\midrule
				\text{Pre-Trained} & 113 & 48.25 & 32.56 & 31.61 & 60.65 & 63.18 & 45.11 & 43.99 & 59.74 & 48.14 \\
				\midrule
				\text{Single-Task}  & 908 & 99.72 & 99.23 & 97.42 & 95.56 & 75.03 & 99.00 & 79.47 & 78.73 & 90.52 \\
				\midrule
				\text{Pruning $\vtheta_t$ } (10\%) &91 & 9.82 & 0.70 & 8.49 & 3.00 & 0.25 & 9.52 & 1.60 & 0.60 & 4.25 \\
				\text{Pruning $\vtheta_t$ } (20\%) &181& 10.28 & 1.77 & 6.71 & 2.11 & 0.28 & 16.11 & 2.45 & 0.46 & 5.02 \\
				\text{Pruning $\vtheta_t$ } (30\%) &271& 9.91 & 3.72 & 17.62 & 2.63 & 0.44 & 10.78 & 2.23 & 0.49 & 5.98\\
				\text{Pruning $\vtheta_t$ } (40\%) &362& 10.09 & 5.08 & 6.29 & 4.48 & 0.32 & 14.48 & 2.71 & 0.40 & 5.48 \\
				\text{Pruning $\vtheta_t$ } (50\%) &452& 10.94 & 5.59 & 20.45 & 6.73 & 0.92 & 17.00 & 7.23 & 0.53 & 8.67\\
				\text{Pruning $\vtheta_t$ } (60\%) &542& 84.54 & 43.72 & 63.88 & 44.63 & 15.23 & 34.67 & 31.91 & 3.47 & 40.26  \\
				\text{Pruning $\vtheta_t$ } (70\%) & 632 & 98.83 & 80.37 & 91.32 & 77.48 & 48.49 & 70.11 & 56.22 & 38.75 & 70.20\\
				\text{Pruning $\vtheta_t$ } (80\%) & 723 & 99.55 & 95.04 & 96.35 & 88.70 & 64.13 & 87.81 & 72.18 & 65.24 & 83.63 \\
				\text{Pruning $\vtheta_t$ } (90\%) & 814 & 99.69 & 99.06 & 97.39 & 95.24 & 73.59 & 98.81 & 79.10 & 77.08 & 89.99\\
				\midrule
				\text{Post-Pruning ${\bf v}_t$} (1\%) & 123 & 58.41 & 40.61 & 39.38 & 67.08 & 66.63 & 56.26 & 48.83 & 63.95 & 55.14 \\
				\text{Post-Pruning ${\bf v}_t$} (5\%) & 159 & 95.82 & 78.61 & 74.35 & 83.67 & 71.60 & 85.81 & 62.39 & 72.73 & 78.12 \\
				\text{Post-Pruning ${\bf v}_t$} (10\%) & 204 & 99.17 & 95.30 & 93.85 & 92.13 & 74.39 & 96.37 & 71.97 &  77.09  & 87.53 \\
				\midrule\rowcolor{Gray}
				\text{BYOM-FFT ${\bf u}_t$ } (1\%) & 123 & 96.17 & 76.33 & 79.27 & 78.03 & 66.88 & 84.89 & 58.03 & 65.99 & 75.70 \\\rowcolor{Gray}
				\text{BYOM-FFT ${\bf u}_t$ } (5\%) & 159 & 99.12 & 92.66 & 91.86 & 88.48 & 71.35 & 94.85 & 67.77 & 73.08 & 84.90 \\\rowcolor{Gray}
				\text{BYOM-FFT ${\bf u}_t$ } (10\%) & 204 &  99.49 &  97.57 & 95.92 & 93.00 & 73.52 & 97.63 & 72.98 & 76.92 & 88.38 \\
				\bottomrule
			\end{NiceTabular}
		} 
	\end{table}
	
	\subsection{T-SNE Visualization for Merging LoRA Finetuned Models}
	\label{sec:vis-lora}
	
\begin{figure*}[!h]
		\centering
		%	\vskip -.2in
		\!\!
		\subfigure[\label{fig:rank_tv}TaskArithmetic.]{\includegraphics[width=0.19\textwidth]{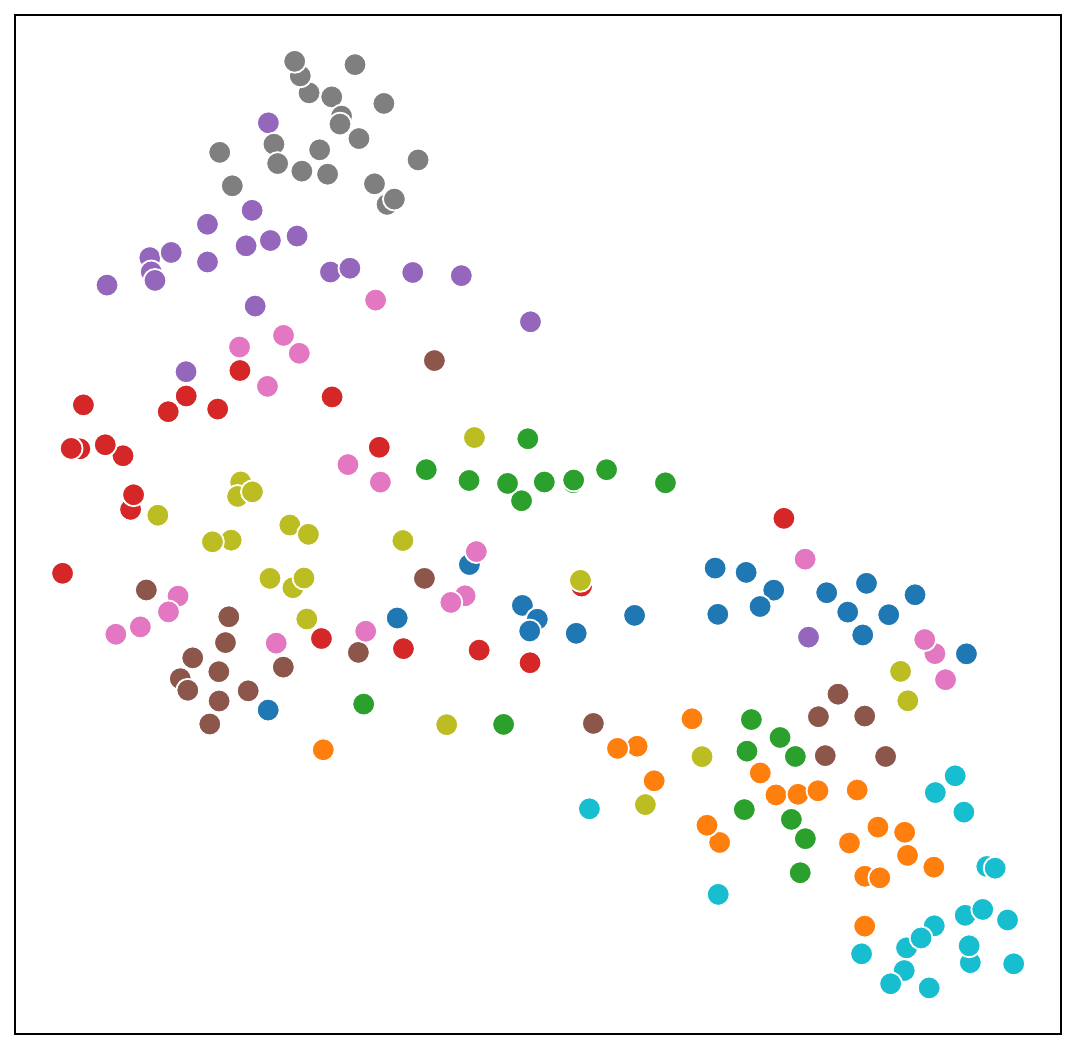}} \!\!
		\subfigure[\label{fig:rank_fm}FisherMerging.]{\includegraphics[width=0.19\textwidth]{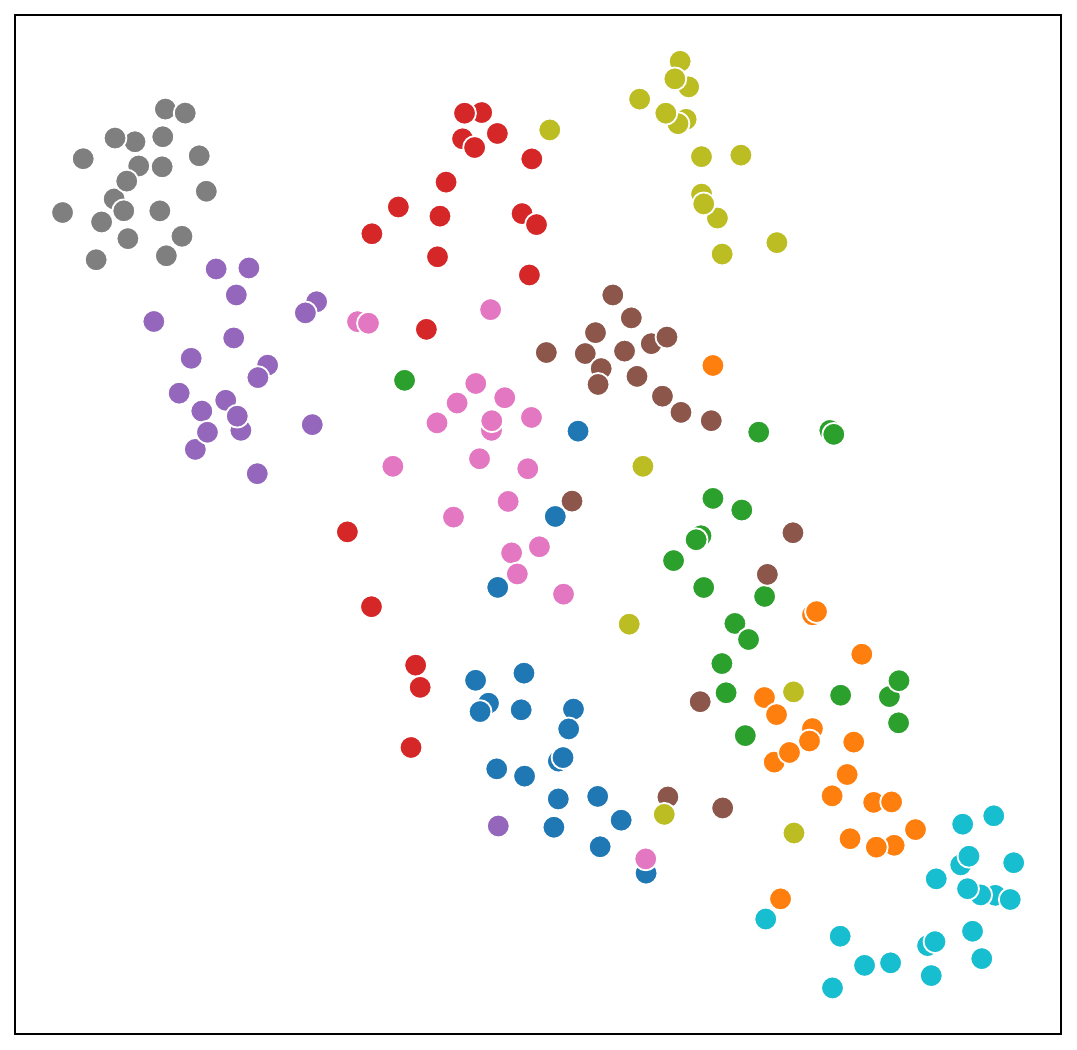}} \!\!
		\subfigure[\label{fig:rank_rm}RegMean.]{\includegraphics[width=0.19\textwidth]{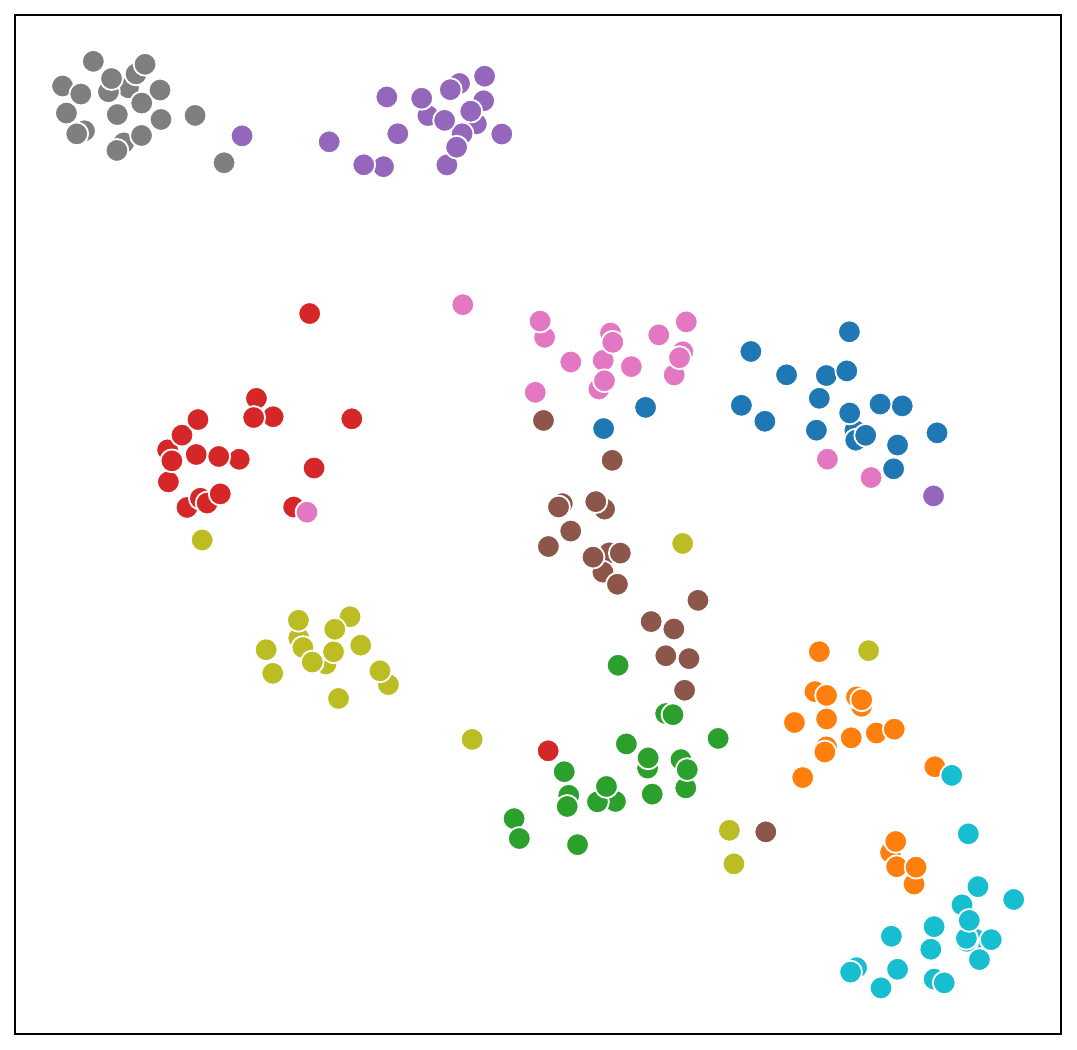}} \!\!  
		\subfigure[\label{fig:rank-tm}TiesMerging.]{\includegraphics[width=0.19\textwidth]{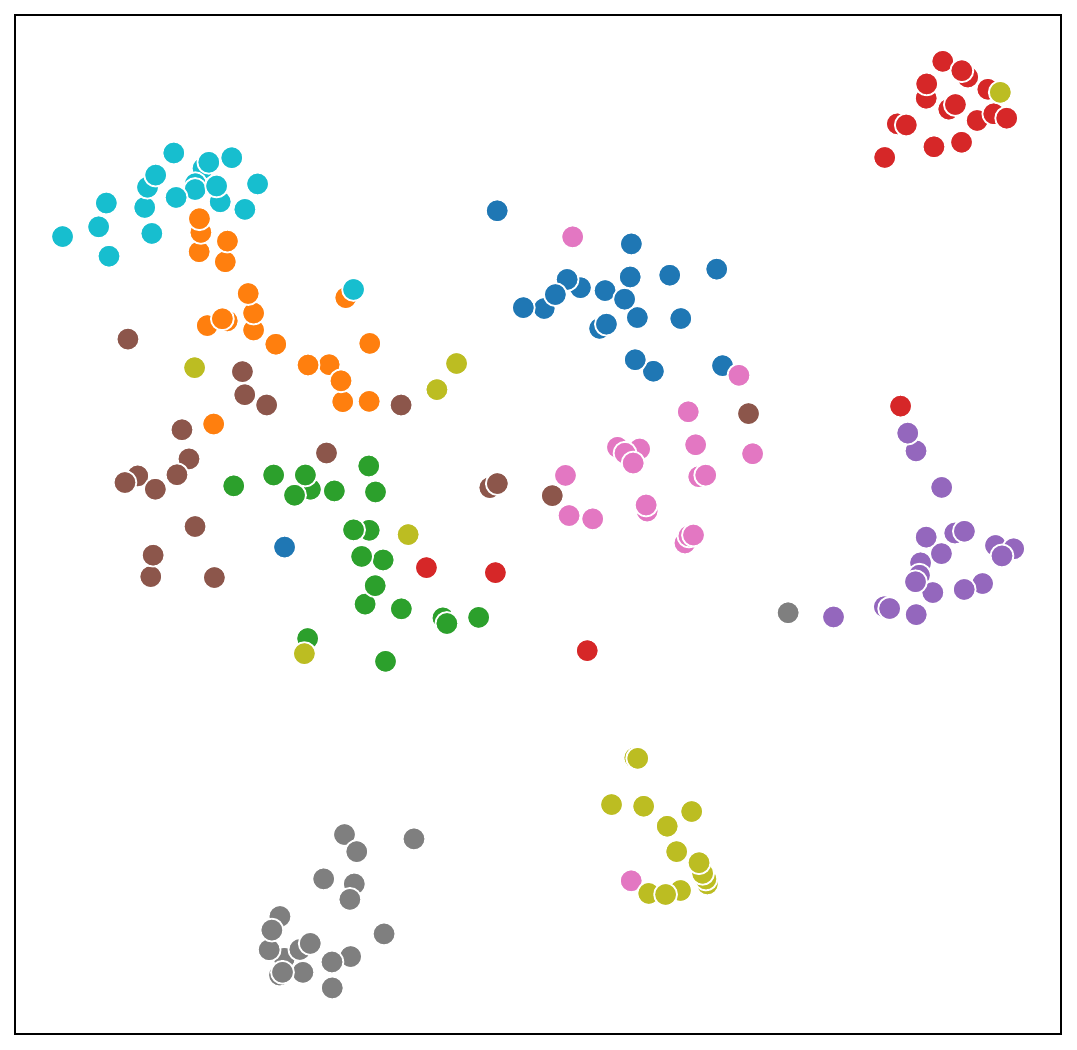}} \!\! 
		\subfigure[\label{fig:rank_init}\!BYOM-LoRA.\!\!\!]{\includegraphics[width=0.19\textwidth]{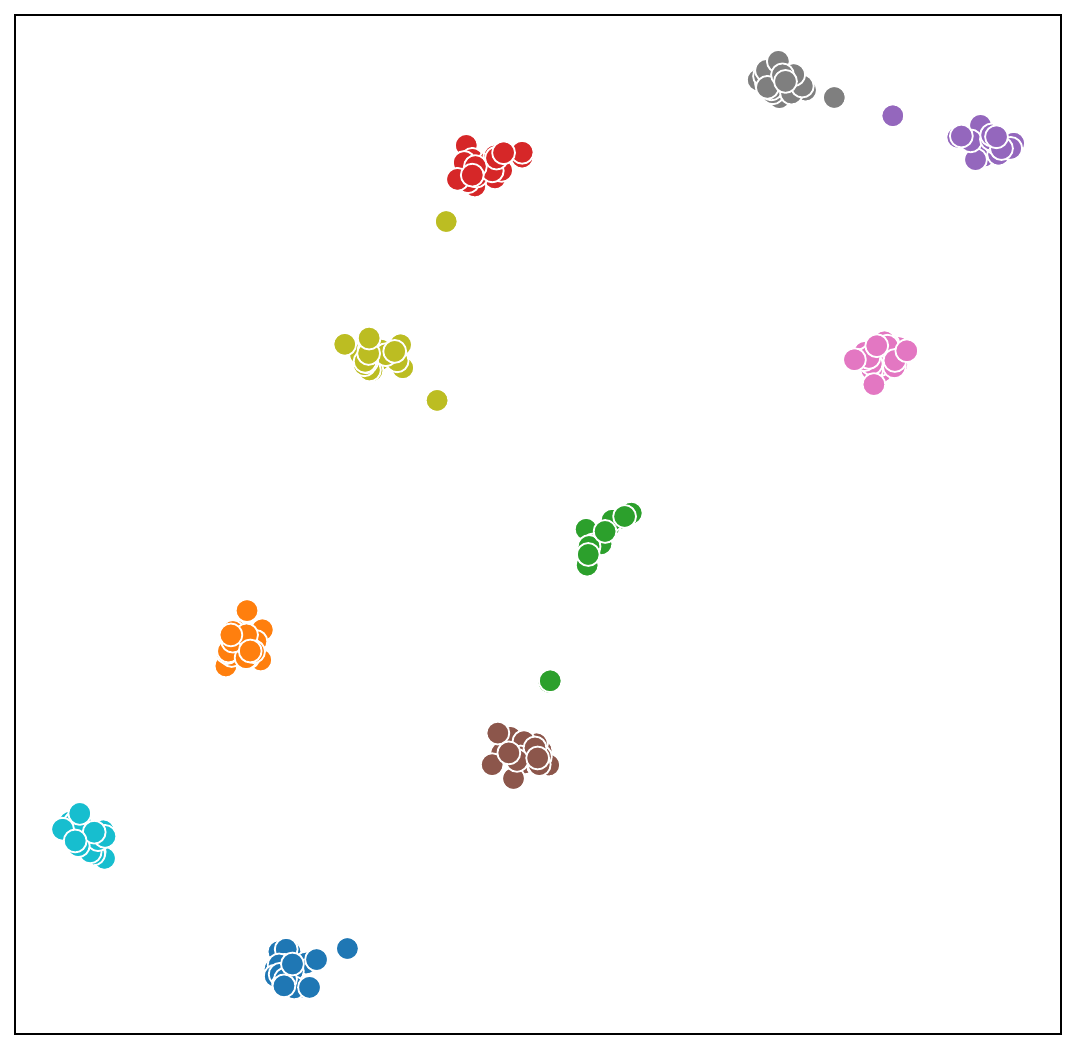}} \!\!
		% \vskip -.2in
		\caption{t-SNE of samples from \textit{EuroSAT} for methods reusing LoRA finetuned \textit{ViT-B/32} Models.}
		\label{fig:lora_tsne}
		% \vskip -.2in
	\end{figure*}

	Figure~\ref{fig:lora_tsne} visualizes the t-SNE of embeddings extracted from 200 images (20 images per class) randomly sampled from \textit{EuroSAT} for methods reusing LoRA finetuned \textit{ViT-B/32} models. 
	As can be seen, BYOM-LoRA (16) has a more compact and separable structure than existing merging methods.

\end{document}